\PassOptionsToPackage{table}{xcolor}
\PassOptionsToPackage{numbers,square,sort&compress,comma}{natbib}
\documentclass[runningheads]{llncs}

\usepackage[preprint]{colm}
\setcitestyle{numbers,square,comma}

\usepackage[T1]{fontenc}
\usepackage[utf8]{inputenc}
\usepackage{lmodern}     
\usepackage{microtype}

\usepackage{amsmath,amssymb,amsfonts}
\usepackage{mathtools}

\usepackage{newtxtext}           
\usepackage[scaled=0.95]{helvet} 

\usepackage{bm}

\AtBeginDocument{\DeclareMathAlphabet{\mathbb}{U}{msb}{m}{n}
}
\usepackage{setspace,enumitem,graphicx,float,booktabs,multirow,array,caption,subcaption,xspace,placeins}

\usepackage{xcolor}
\definecolor{citeblue}{HTML}{2b5bca}       
\definecolor{linkred}{HTML}{9e1316}        
\definecolor{boxbackground}{HTML}{F3F4F7}  
\definecolor{bestcolor}{RGB}{255,224,138}  
\definecolor{secondcolor}{RGB}{208,231,255}
\definecolor{zeyucolor}{RGB}{218,165,32}   

\usepackage[ruled,vlined,linesnumbered]{algorithm2e}
\usepackage[many]{tcolorbox}
\usepackage{tikz}
\usetikzlibrary{arrows.meta, positioning, shapes.geometric, fit, backgrounds, calc, decorations.pathreplacing, patterns}

\newtcolorbox{abstractbox}{
    colback=boxbackground,  
    colframe=boxbackground, 
    boxrule=0pt,            
    arc=4mm,                
    auto outer arc,
    left=18pt, right=18pt, top=18pt, bottom=18pt,            
    width=\linewidth,       
    halign=justify          
}

\usepackage{hyperref} 
\hypersetup{
    colorlinks=true,
    linkcolor=citeblue,
    citecolor=citeblue,  
    urlcolor=citeblue,
    bookmarksnumbered=true
}

\usepackage{fancyhdr}
\renewcommand{\headrulewidth}{1pt}
\usepackage{tocloft}

\cftsetindents{section}{0em}{1.5em}
\cftsetindents{subsection}{1.5em}{2.3em}
\cftsetindents{subsubsection}{3.8em}{3.2em}
\newcommand{\HyVLA}{\texttt{HyVLA-}\texttt{0}.\texttt{5}\xspace}

\providecommand{\eg}{\emph{e.g.}\xspace}
\providecommand{\ie}{\emph{i.e.}\xspace}

\newcommand{\paragrapha}[1]{\par\addvspace{0.45em}\noindent\textbf{#1}}

\providecommand{\loss}{\mathcal{L}}
\providecommand{\dataset}{\mathcal{D}}
\providecommand{\policy}{\pi_\theta}
\providecommand{\refpolicy}{\pi_{\text{ref}}}
\providecommand{\expect}{\mathbb{E}}

\begin{document}


\makeatletter
\def\@makefnmark{}
\makeatother

\begin{abstractbox}

{\sffamily\bfseries\Large Hy-Embodied-0.5-VLA: From Vision-Language-Action Models to a Real-World Robot Learning Stack
 \par}
\vspace{0.6em} 

{\sffamily\textbf{Tencent Robotics X}~~~~~~\textbf{Futian Laboratory}~~~~~~\textbf{Tencent Hy Team}\par}
\vspace{1.5em} 

The past year has witnessed a rapid proliferation of Vision-Language-Action (VLA) models, with growing attention now turning to the next generation of embodied foundation models. However, a truly generalist robot is unlikely to emerge from any single model in isolation. Rather, it must be built on a full robot learning stack that remains robust from data collection to real-world deployment. 

In this report, we present Hy-Embodied-0.5-VLA, abbreviated as \textbf{\HyVLA}, an end-to-end system that spans the full robot learning stack: data collection, model design, continued pre-training and supervised fine-tuning, RL post-training, and real-world deployment. Each component serves a distinct role in this stack. \textbf{For data}, we develop a custom fingertip UMI device with a motion-capture cage to collect over 10,000 hours of egocentric, sub-millimeter-precision human demonstrations that can also directly serve as post-training trajectories. \textbf{For modeling}, we extend the Hy-Embodied-0.5 backbone with a flow-matching action expert, a compact memory encoder, and a delta-chunk action representation that decouples policy learning from embodiment-specific kinematics. \textbf{For continued pre-training and fine-tuning}, starting from the checkpoint continued-pre-trained on the UMI corpus, we introduce two real-robot SFT tracks: Track-A for target-robot adaptation and Track-B for UMI-only cross-embodiment transfer. \textbf{For RL post-training}, we introduce a Proximalized Preference Optimization (PRO)-based offline RL algorithm that turns failure cases into rapid policy improvement and drives performance toward near-ceiling success rates without requiring a learned reward model. \textbf{For deployment}, an asynchronous inference pipeline with lightweight trajectory smoothing enables high-frequency closed-loop control. Taken as a whole, the \texttt{Hy-Embodied-0.5-VLA} stack marks a meaningful step toward deployable generalist robots.

\vspace{1.5em} 

\noindent\textbf{Website:} \href{https://tairos.tencent.com/openSourceModels/hy-embodied-0.5-vla}{{\small\nolinkurl{tairos.tencent.com/openSourceModels/hy-embodied-0.5-vla}}}\\
\textbf{Github:} \href{https://github.com/Tencent-Hunyuan/Hy-Embodied-0.5-VLA}{{\small\nolinkurl{github.com/Tencent-Hunyuan/Hy-Embodied-0.5-VLA}}}\\
\textbf{Model:} \href{https://huggingface.co/tencent/Hy-Embodied-0.5-VLA-UMI}{{\small\nolinkurl{huggingface.co/tencent/Hy-Embodied-0.5-VLA-UMI}}}\\
\textbf{Dataset:} \href{https://huggingface.co/datasets/tencent/Hy-Embodied-0.5-VLA-Data}{{\small\nolinkurl{huggingface.co/datasets/tencent/Hy-Embodied-0.5-VLA-Data}}}
\end{abstractbox}


\section{Introduction}

\begin{figure}[htbp]
  \centering
  \includegraphics[width=\textwidth]{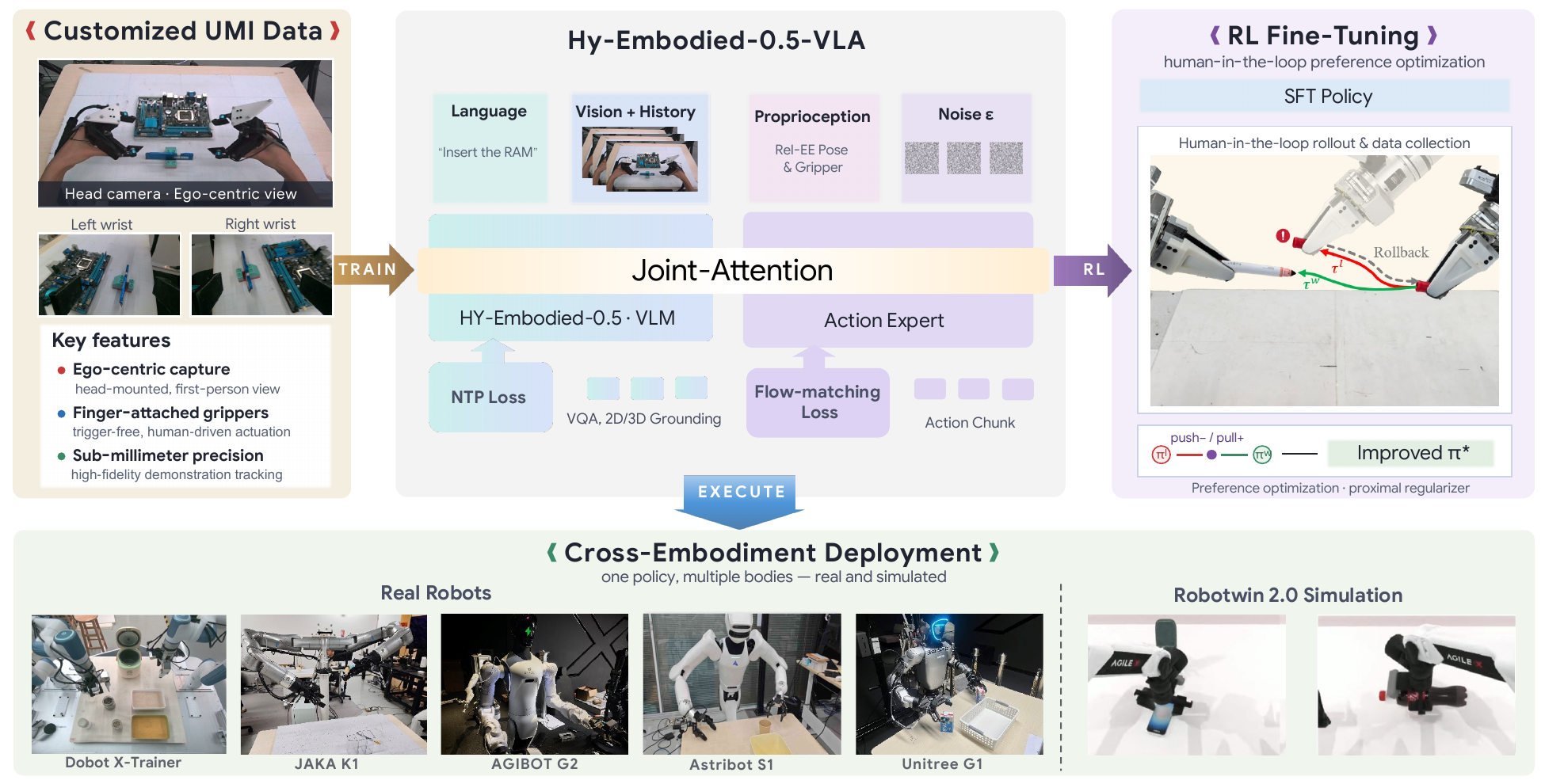}
  \caption{\textbf{Overview of Hy-Embodied-0.5-VLA.} An end-to-end VLA system that pairs the Hy-Embodied-0.5-MoT backbone with a flow-matching action expert under a delta-chunk action representation, pre-trained on a $10$K-hour egocentric UMI corpus and refined with a reward-free, Proximalized Preference Optimization (PRO)-based offline RL stage (\textit{FlowPRO}). A single pre-trained checkpoint specializes along two parallel post-training tracks for cross-embodiment transfer to morphologically unseen robots.}
  \label{fig:teaser}
  \vspace{-4mm}
\end{figure}

Recent advances in Vision-Language-Action (VLA) architectures have demonstrated promising capabilities in continuous robotic control~\citep{black2024pi0,intelligence2025pi05,team2025gemini,bjorck2025gr00t, lin2026universal, liu2025rdt}. Yet turning these model advances into deployable generalist robots requires more than stronger policies: the data, training, adaptation, and execution layers must be co-designed around real-hardware constraints.

These system-level requirements expose three coupled challenges on the data side. First, traditional teleoperation~\cite{zhao2023aloha,zhao2023learning} relies on master--slave interfaces that force operators to unnaturally adapt to the robot's workspace, lacks direct haptic feedback, and therefore precludes delicate manipulation. Second, while leveraging human data~\cite{yang2025egovla,li2025scalable} or hand-held frameworks such as UMI~\cite{chi2024umi} alleviates data scarcity, these alternatives introduce new limitations: raw human demonstrations greatly enrich behavioral diversity but provide overly coarse action labels, and existing UMI rigs improve localization through SLAM at the cost of cumbersome handheld devices that fail to capture fingertip-level force transmission. Third, bridging the cross-embodiment gap involves more than adapting kinematics: it requires addressing the embodiment gap between human and robot motion spaces, the control gap induced by different dynamics and actuation, and the perception gap between human egocentric views and robot-mounted camera observations. 

Beyond data, the architectural design, training paradigms, and deployment stack of VLA models present equally critical bottlenecks. Early approaches largely relied on autoregressive modeling over discretized action tokens~\citep{brohan2023rt2,kim2024openvla}, which inherently limits both execution speed and control precision. Recent frameworks mitigate this by coupling a Vision-Language Model with a flow-matching action expert that predicts continuous actions~\citep{black2024pi0}, yet their foundational visual backbones are not explicitly engineered for robotic control: a significant gap remains between generalist visual representations and the dense spatiotemporal reasoning required for physical interaction. On top of representational issues, standard imitation learning struggles to reach last-mile dexterity, while existing reinforcement learning recipes for continuous control typically depend on brittle reward models or value networks~\citep{intelligence2025pi06}. Finally, even a well-trained policy is of limited use unless it can be served at high frequency in a closed visual loop on real hardware---a deployment constraint that is rarely treated as a first-class design target. Addressing these combined bottlenecks therefore requires a unified pipeline that jointly tackles data, model, policy refinement, and deployment.

To address these challenges, we present \texttt{Hy-Embodied-0.5-VLA} (Fig.~\ref{fig:teaser}), an end-to-end system that spans the full stack---from custom data-collection hardware to production-ready deployment. Rather than treating VLA modeling as an isolated problem, \HyVLA is organized as a complete pipeline in which data, modeling, RL post-training, and deployment each serve a distinct role.

\textbf{For data}, we build a custom fingertip UMI device paired with a motion-capture cage, and use it to collect over $10$K hours of egocentric, sub-millimeter-precision human demonstrations. The fingertip form factor restores natural haptic perception that bulky handheld rigs cannot offer; the motion-capture cage produces high-fidelity action labels beyond the reach of SLAM-only pipelines; and the egocentric viewpoint supplies global semantic context rather than over-relying on local wrist cameras. Crucially, the same trajectories can also directly serve as post-training data, making them reusable for downstream adaptation and reducing the need for separate target-robot data collection~(Sec.~\ref{sec:umi-corpus}).

\textbf{For modeling}, we extend our Hy-Embodied-0.5~\cite{hyembodied2025} backbone---a 4B Mixture-of-Transformers VLM pre-trained on embodied corpora---with a flow-matching action expert for continuous, high-frequency action prediction. Compared with adapting general-purpose VLMs~\cite{beyer2024paligemma,wang2024qwen2,li2025eagle}, this embodied-native initialization yields stronger spatial priors and faster post-training convergence. We further introduce a compact memory encoder for spatiotemporal context, and adopt a delta-chunk action representation that predicts incremental end-effector motion between consecutive steps. The delta-chunk formulation decouples policy learning from embodiment-specific kinematics and substantially shrinks the optimization search space, providing a clean substrate for cross-embodiment post-training and deployment~
(Secs.~\ref{sec:model} and \ref{sec:deployment}).

\textbf{For continued pre-training and fine-tuning}, we first pre-train \HyVLA on the $10$K-hour UMI corpus, then specialize the resulting checkpoint through task-specific supervised fine-tuning. Real-robot SFT is organized into two tracks: Track-A studies intra-embodiment adaptation with target-robot demonstrations and deployment on the same platform, while Track-B studies UMI-only cross-embodiment transfer to morphologically different robots without target-robot teleoperation~(Sec.~\ref{sec:stage2}).

\textbf{For RL post-training}, we introduce \textit{FlowPRO}~\cite{wu2026flowpro}, a critic-free, reward-free Proximalized Preference Optimization (PRO)-based offline reinforcement learning algorithm. Through a teleoperated intervention-and-rollback pipeline, paired success/failure trajectories are harvested directly from policy rollouts. An RPRO loss then aligns these preferences with the continuous flow-matching objective, while a contrastive gradient-cancellation property suppresses catastrophic forgetting. FlowPRO turns failure cases into a rapid iteration loop for improving long-tail manipulation robustness and driving performance toward near-ceiling success rates, without training any reward or value network~(Sec.~\ref{sec:rl_posttraining}).

\textbf{For deployment}, we implement an asynchronous inference framework that overlaps backbone forward passes with action execution, and stitches successive delta chunks via a simple yet effective cubic B\'ezier action smoother that guarantees $C^1$-continuous transitions~(Sec.~\ref{sec:inference}). Together, these components enable high-frequency, closed-loop control on real hardware and complete the path from data collection to real-world operation on the factory floor. The rest of this report details how the full \HyVLA pipeline is built, trained, and validated across large-volume pre-training, cross-embodiment post-training, PRO-based refinement, and physical robot deployment.

\section{Model Architecture}
\label{sec:model}

\begin{figure}[!htbp]
  \centering
  \includegraphics[width=\textwidth]{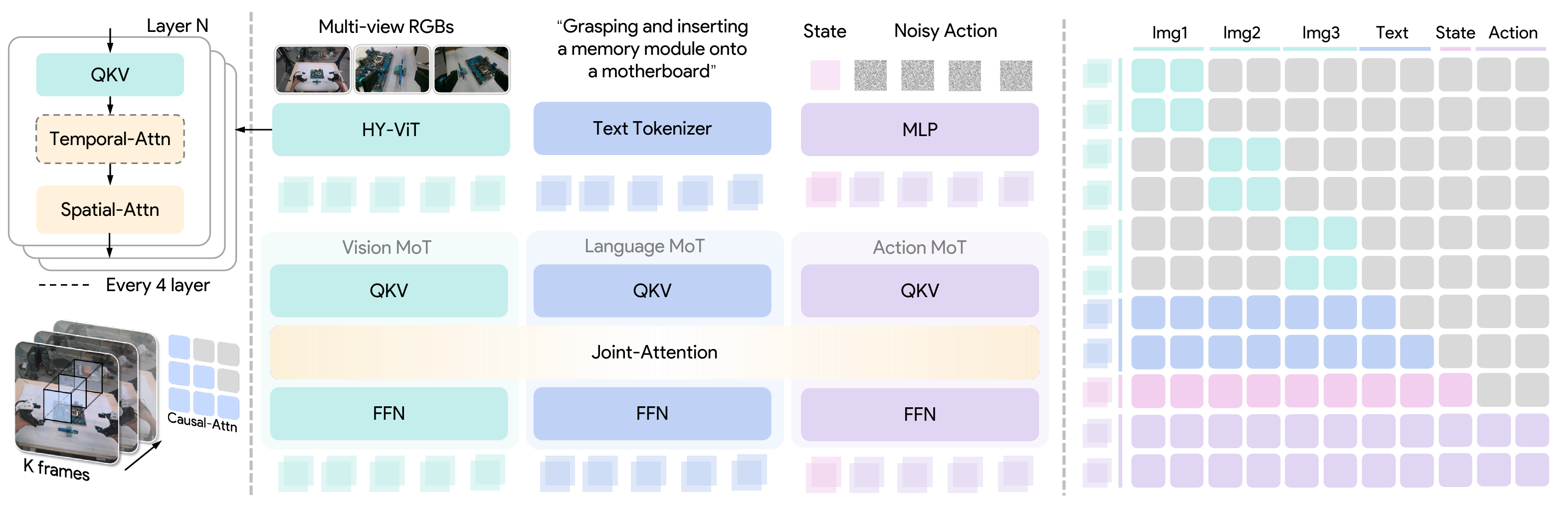}
\caption{\textbf{Architectural overview of \HyVLA.} The framework adopts a MoT architecture to facilitate cross-modal interactions via a shared joint-attention mechanism. To effectively process $K$-frame multi-view RGB sequences, the image encoder is extended into a compact memory encoder. Specifically, temporal attention blocks are interleaved every four layers to enforce causal masking across the temporal dimension and seamlessly incorporate historical visual context. As depicted on the right, the attention mask demonstrates our block-wise causal attention strategy. Following Hy-Embodied-0.5~\citep{hyembodied2025}, we apply local bidirectional attention to model the multi-view observations.}

\label{fig:hyvla_arch}

\end{figure}

\HyVLA follows the vision-language-action (VLA) paradigm, in which a pre-trained vision-language model (VLM) supplies broad semantic perception and a dedicated action module translates the resulting multi-modal context into low-level robot control (Fig.~\ref{fig:hyvla_arch}). On top of this paradigm, \HyVLA comprises three components. \textit{Firstly}, the backbone is the embodied VLM Hy-Embodied-0.5~\citep{hyembodied2025}, which adopts a Mixture-of-Transformers (MoT) architecture~\citep{liang2024mot} with modality-adaptive computation and native-resolution image encoding. \textit{Secondly}, an action expert generates continuous action chunks through conditional flow matching~\citep{black2024pi0,lipman2023fm}, with the robotics-specific state and action streams kept separate from the VLM and coupled to it through shared attention. \textit{Finally}, the image encoder is extended into a compact memory encoder that aggregates a multi-frame observation history through interleaved temporal-spatial attention~\citep{intelligence2026mem}. We first formalize the problem in Sec.~\ref{sec:formulation}, and then detail the backbone (Sec.~\ref{sec:backbone-arch}), the action expert (Sec.~\ref{sec:action-expert}), and the compact memory encoder (Sec.~\ref{sec:short-term-mem}).

\subsection{Problem Formulation}
\label{sec:formulation}
We formulate manipulation as a goal-conditioned, chunk-level control problem. At every decision step $t$, the policy consumes a multi-modal observation $\mathbf{o}_t$ and predicts a chunk of future actions $\mathbf{A}_t$; that is, we model the conditional distribution $p(\mathbf{A}_t \mid \mathbf{o}_t)$. Formally,
\begin{equation}
  \mathbf{o}_t = \big(\mathbf{I}_t,\ \ell,\ \mathbf{s}_t\big),
  \quad
  \mathcal{I}_t = \big\{\, \mathbf{I}^{\,v}_{\,t-k} \,\big\}_{\,v=1:n}^{\,k=0:K-1},
  \qquad
  \mathbf{A}_t = \big( \mathbf{a}_t,\ \mathbf{a}_{t+1},\ \dots,\ \mathbf{a}_{t+H-1} \big),
  \label{eq:io}
\end{equation}
where $\mathcal{I}_t$ is the visual stream, $\ell$ the language instruction, $\mathbf{s}_t$ the proprioceptive state, and $A_t$ the predicted action chunk of horizon $H$. We describe each component below.

\paragrapha{Visual Input.} The visual stream $\mathcal{I}_t$ is a \emph{multi-view, multi-frame} RGB observation: at step $t$ it comprises the $K$ most recent frames from each of the $n$ camera viewpoints (\eg a head-mounted view together with a wrist-mounted view per arm), \ie $n{\times}K$ images in total. The history length $K$ is a configurable hyperparameter; its value at each training stage is specified in Sec.~\ref{sec:training}, with $K{=}1$ recovering the single-frame case.

\paragrapha{Language Input.} A natural-language task instruction $\ell$ (\eg \texttt{``hang the mug on the rack''}) defines the goal. It is tokenized and jointly encoded with the visual stream by the VLM backbone, enabling the policy to ground its behaviors in the commanded semantics.

\paragrapha{Proprioceptive Input.} The robot state $\mathbf{s}_t$ encodes the current pose of the controlled end-effector(s) and is projected into the backbone embedding space, providing the embodiment-grounded context that anchors action prediction to the robot's present configuration.

\paragrapha{Action Output.} Instead of single-step execution, the policy predicts an entire action chunk~\citep{zhao2023aloha} of horizon $H$ per inference cycle. This ensures temporally smooth, high-frequency control while significantly reducing the inference latency, as the VLM backbone is evaluated only once to condition the entire $H$-step generation via flow matching.


\paragrapha{End-effector-frame Representation.} Both the proprioceptive state $s_t$ and the action $\mathbf{a}_{t'}$ are formulated in the \emph{end-effector frame} (EEF), an embodiment-agnostic representation that decouples the policy from robot-specific joint kinematics. For each controlled arm, a pose is parameterized by a $3$-D Cartesian translation (\texttt{xyz}) and a $6$-D continuous rotation representation~\citep{zhang2018mode}, augmented by a $1$-D normalized gripper command, \ie, $\mathbf{s}_t, \mathbf{a}_{t'} \in \mathbb{R}^{10}$ per arm. The proprioceptive state $s_t$ is defined in the end-effector frame with respect to the embodiment root, while each future action $\mathbf{a}_{t'}$ is a delta-chunk defined in the \emph{relative EEF} that takes the current state $s_t$ as its reference frame.

\paragrapha{Optional Co-Training Tasks.} Beyond learning from action-labeled trajectories, the unified VLA architecture integrates auxiliary next-token prediction tasks to preserve its foundational vision-language reasoning and spatial grounding capabilities. We denote this auxiliary data mixture as $\mathcal{D}_{\mathrm{ct}} = \mathcal{D}_{\mathrm{VQA}} \cup \mathcal{D}_{\mathrm{2D}} \cup \mathcal{D}_{\mathrm{3D}}$. Each training instance is formulated as a pair $(\mathbf{c}, y_{1:M})$, where $\mathbf{c}$ represents the vision-language conditions, and $y_{1:M}$ denotes a sequence of $M$ serialized target tokens. Depending on the specific task, $y_{1:M}$ consists of semantic answer tokens for VQA, normalized 2D spatial coordinates, or 3D geometric parameters formulated within the camera or scene frame. 
Crucially, this co-training objective directly optimizes the parameters of the shared VLM backbone, ensuring it maintains and enriches the vital semantic and spatial representations.


\subsection{Hy-Embodied: Modality-Adaptive Computing Backbone}
\label{sec:backbone-arch}
\HyVLA builds upon the embodied VLM \textbf{Hy-Embodied-0.5-MoT}~\citep{hyembodied2025}, a compact model with $4$\,B parameters optimized for edge deployment. It instantiates the standard image-encoder-plus-language-model recipe, and we detail three key design choices adapted for manipulation.

\paragrapha{Native-resolution visual encoding.} The backbone encodes images with Hy-ViT~2.0, a native-resolution Vision Transformer (ViT)~\cite{dosovitskiy2020image, dehghani2023patch} that accepts arbitrary input resolutions and is distilled from a larger internal teacher. Each camera stream can therefore be processed at its native resolution rather than being down-sampled to a fixed size.

\paragrapha{Modality-adaptive Computation via MoT.} The backbone adopts a Mixture of Transformers (MoT) architecture~\citep{liang2024mot}, which is directly initialized with the pre-trained weights of HY-Embodied-0.5~\cite{hyembodied2025}. This design maintains \emph{non-shared} QKV and FFN parameters for the visual and textual streams. Specifically, during the forward pass, all visual tokens extracted by the ViT are computed using a duplicated, vision-specific parameter set, whereas textual tokens are processed using the original language parameters. Cross-modal interaction is strictly limited to the shared self-attention layers. Consequently, the visual and textual parameters are updated independently. Furthermore, following the original configuration of HY-Embodied-0.5~\cite{hyembodied2025}, the backbone applies bidirectional attention strictly \emph{among} the visual tokens of each individual image, while maintaining standard causal attention for the language tokens.



\paragrapha{Co-training Objective.} For the auxiliary VQA and spatial grounding instances sampled from $\mathcal{D}_{\mathrm{ct}}$, the VLM backbone employs its native language modeling head to autoregressively decode the serialized target tokens. We optimize this process via a standard next-token prediction objective:
\begin{equation}
  \mathcal{L}_{\text{ntp}}(\theta) =
  \mathbb{E}_{(\mathbf{c},y)\sim\mathcal{D}_{\mathrm{ct}}}
  \left[ \, -\sum_{j=1}^{M} \log p_\theta\big(y_j \mid \mathbf{c}, y_{<j}\big) \, \right],
  \label{eq:ntp}
\end{equation}
where $y_j$ denotes the $j$-th serialized target token.


\subsection{Action Expert with Dual-Tower Flow Matching}
\label{sec:action-expert}
Rather than discretizing actions into language-like tokens, \HyVLA equips the backbone with an \emph{action expert} that models the continuous distribution $p(\mathbf{A}_t \mid \mathbf{o}_t)$ directly via conditional flow matching~\citep{lipman2023fm}.

\paragrapha{Dual-tower Routing.} On top of the MoT backbone, \HyVLA separates the joint transformer into an understanding-oriented VLM tower and a generation-oriented action-expert tower. The VLM tower processes visual and textual context with the modality-adaptive parameters described above, while the action expert consumes the projected robot state and noisy action tokens $[\mathbf{s}_t, \mathbf{A}_t^\tau]$ to produce the continuous action velocity field. The two towers interact through shared self-attention, allowing grounded visual-language context to guide action generation.

\paragrapha{Block-wise Causal Attention.} We partition the token sequence into three blocks, $[\,\mathcal{I}_t, \ell\,]$, $[\,\mathbf{s}_t\,]$, and $[\,\mathbf{a}^\tau_{t,0}, \dots, \mathbf{a}^\tau_{t,H-1}\,]$, and apply attention that is bidirectional \emph{within} each block but strictly causal \emph{across} blocks. The perception block is prevented from attending to the robotics-specific blocks, minimizing distribution shift from VLM pre-training; the state block is isolated so that its keys and values can be cached; and the noisy-action block attends to the full prefix.

\paragrapha{Flow-matching Objective.} Let $\mathbf{A}^\tau_t = \tau \mathbf{A}_t + (1-\tau)\epsilon$ with $\epsilon \sim \mathcal{N}(\mathbf{0}, \mathbf{I})$ denote the noisy action chunk at flow timestep $\tau \in [0,1]$. The action expert regresses the velocity field $v_\theta$ that transports noise to the target actions, trained with
\begin{equation}
  \mathcal{L}_\text{fm}(\theta) \;=\; \mathbb{E}_{\,p(\mathbf{A}_t \mid \mathbf{o}_t),\, q(\mathbf{A}^\tau_t \mid \mathbf{A}_t)}
  \big\| v_\theta(\mathbf{A}^\tau_t, \mathbf{o}_t) - (\epsilon - \mathbf{A}_t) \big\|_2^2 ,
  \label{eq:fm}
\end{equation}
where $\mathbf{A}_t$ is the ground-truth chunk, $\mathbf{A}^\tau_t$ its noised version, $\mathbf{A}_\theta(\mathbf{A}^\tau_t, \mathbf{o}_t)$ the predicted velocity conditioned on the observation $o_t$, and $\epsilon - A_t$ the target denoising direction. The flow timestep $\tau$ is sampled from a Beta distribution skewed toward high-noise regimes, which emphasizes the harder, more informative stages of action denoising. When auxiliary data is mixed with robot demonstrations, the total objective is $\mathcal{L}(\theta)=\mathcal{L}_{\text{fm}}(\theta)+\lambda_{\text{ntp}}\mathcal{L}_{\text{ntp}}(\theta)$, with $\lambda_{\text{ntp}}{=}0$ recovering action-only training.

\paragrapha{Inference.} At deployment, the policy generates an action chunk by integrating the learned velocity field from $\tau{=}0$ to $\tau{=}1$ via the forward Euler update $\mathbf{A}^{\tau+\delta}_t = \mathbf{A}^\tau_t + \delta\, \mathbf{v}_\theta(\mathbf{A}^\tau_t, \mathbf{o}_t)$ over $10$ integration steps ($\delta{=}0.1$). Because the conditioning observation prefix $\mathbf{o}_t$ remains constant across all solver iterations, its keys and values are cached during the initial forward pass. Consequently, subsequent steps exclusively recompute the action tokens, significantly reducing computational overhead.


\subsection{Compact Memory Encoder with Temporal-Spatial Attention}
\label{sec:short-term-mem}
\HyVLA conditions on the $K$-frame multi-view history $\mathcal{I}_t$ of Eq.~\eqref{eq:io} to form a compact memory encoding. Encoding all $n{\times}K$ frames independently and forwarding them to the backbone would multiply the visual token count passed to the VLM. We instead extend the image encoder into a \emph{video encoder} that compresses the temporal dimension before tokens reach the VLM backbone.

\paragrapha{Factorized Temporal-spatial Attention.} Following Pi-MEM~\citep{intelligence2026mem}, the video encoder preserves the patchify-then-attend structure of a standard ViT and inserts a temporal pass once every $L$ layers. At such a layer, we add a fixed sinusoidal temporal encoding $e(k)$ (with $e(0)=\mathbf{0}$) and reuse the \emph{same} $\mathrm{QKV}$ and output projection $W_O$ of the underlying ViT block, then factorize the attention into two passes that share these projections:
\begin{align}
  \text{(temporal)}\quad &\tilde{\mathbf{V}}_p = \mathrm{CausalAttn}\big(\mathbf{Q}_p, \mathbf{K}_p, \mathbf{V}_p\big), && \text{over the $K$ frames at each patch $p$;} \label{eq:mem-time}\\
  \text{(spatial)}\quad &\tilde{\mathbf{X}}_k = \mathbf{W}_O\,\mathrm{Attn}\big(\mathbf{Q}_k, \mathbf{K}_k, \tilde{\mathbf{V}}_k\big), && \text{over the $n$ patches within each frame $k$,} \label{eq:mem-space}
\end{align}
where $\tilde X_k$ is the attention output that the block feeds into its residual connection and MLP. The temporal pass is a causal attention, so each frame attends only to the present and past, matching the streaming nature of on-robot perception. The spatial pass is the original bidirectional self-attention within a frame, applied to the time-mixed values $\tilde V$. This factorization avoids the $\mathcal{O}(n^2K^2)$ cost of joint space-time attention and reduces the per-layer cost to $\mathcal{O}(Kn^2 + nK^2)$.

\paragrapha{Token-count-preserving compression.} In the upper layers of the video encoder we discard the patch representations of past frames and forward only the current-frame tokens to the backbone. Because the interleaved temporal attention has already \emph{baked} the historical context into the current-frame representation, the number of visual tokens passed to the VLM matches that of a single-frame policy.

\paragrapha{Parameter-free, Transfer-friendly Design.} The video encoder introduces \emph{no} new learnable parameters relative to the single-image Hy-ViT~2.0: both passes reuse the $\mathrm{QKV}$ and $W_O$ projections of Eq.~\eqref{eq:mem-time}--\eqref{eq:mem-space}, and the temporal encoding $e(k)$ is a fixed sinusoid with $e(0)=\mathbf{0}$ rather than a learned table. Consequently, when $K{=}1$ the causal temporal attention is the identity and $e(0)=\mathbf{0}$ leaves the input unchanged, so each augmented block reduces \emph{exactly} to the pre-trained ViT block. The memory-augmented backbone is therefore initialized directly from the Hy-Embodied-0.5 weights and recovers the single-frame encoder as a special case.

\section{Pre-training and Supervised Fine-tuning}
\label{sec:training}
This section focuses on the supervised stages of \HyVLA training: large-scale \emph{pre-training} on the Hy-UMI-10K corpus to learn a generalist action prior, followed by \emph{supervised fine-tuning} (SFT) on task-specific demonstrations from each target embodiment.

\subsection{Hy-UMI-10K: High-Fidelity Manipulation Dataset}
\label{sec:umi-corpus}
\begin{figure}[t]
  \centering
  \includegraphics[width=1.0\linewidth]{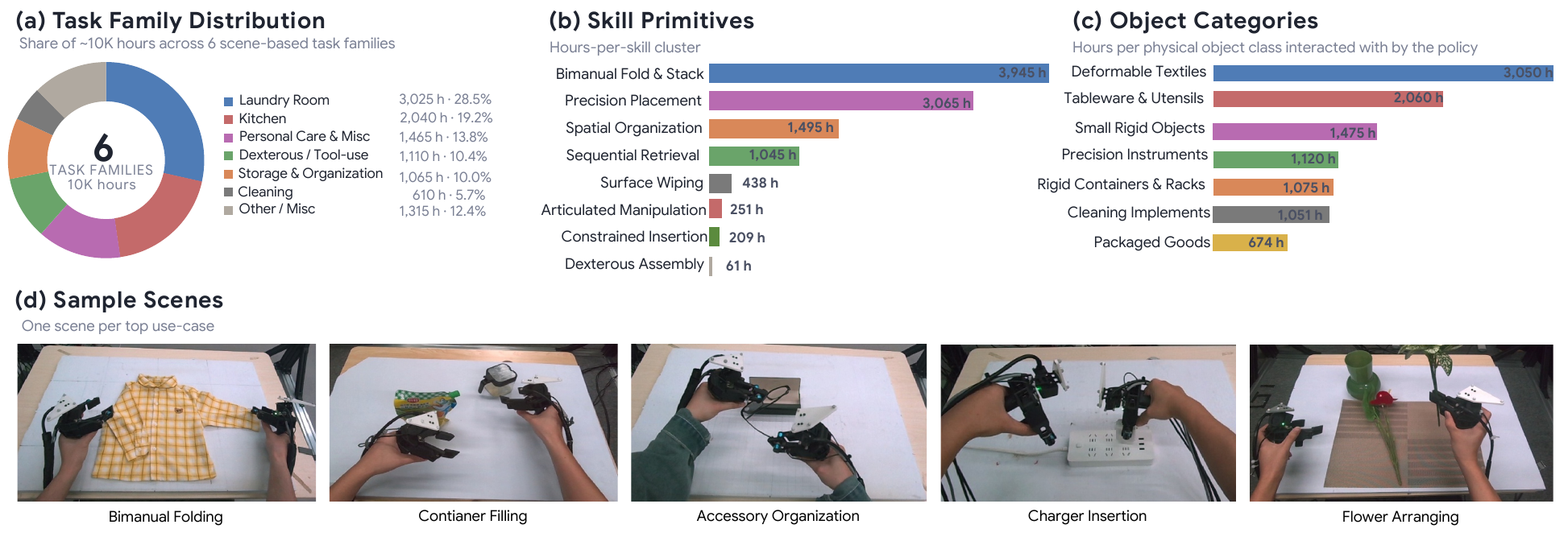}
  \caption{\textbf{UMI dataset distribution.} Detailed characterization of our diverse, in-house collected 10K-hour UMI demonstration corpus. The distribution outlines broad scale, diverse skill categories, environmental conditions, and manipulated objects, ensuring generalist-level manipulation capacity.}
  \label{fig:umi_data_dist}
\end{figure}

\HyVLA is pre-trained on Hy-UMI-10K, a hand-held Universal Manipulation Interface (UMI) dataset~\citep{chi2024umi} of more than $10$K hours collected in-house (Fig.~\ref{fig:umi_data_dist}), and it is the sole data source for pre-training. Unlike standard UMI pipelines that recover gripper poses from on-board visual SLAM, our capture rig tracks each gripper with an external \emph{optical motion-capture} system, which labels every 6-DoF trajectory at sub-millimetre precision in a single, globally consistent world frame---hence \emph{high-fidelity}. We describe its capture device, composition, and the pre-training recipe below.

\begin{figure}[t]
  \centering
  \includegraphics[width=\linewidth]{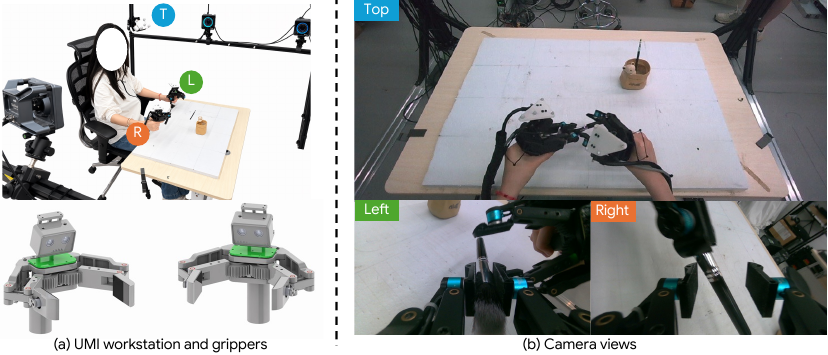}
  \caption{\textbf{UMI custom data collection workstation.} The in-house designed hardware setup features an external optical motion-capture system delivering sub-millimeter high-precision tracking, an ego-centric visual perspective camera with native depth capture, a 6-dimensional force-sensing gripper on each hand.}
  \label{fig:umi_station_design}
\end{figure}

\paragraph{Capture device.} Demonstrations are acquired with custom-designed \emph{hand-held UMI grippers} detached from kinematics of specific embodiments (Fig.~\ref{fig:umi_station_design}). The gripper design follows that of a commonly adopted industrial gripper, Changingtek CTAG2F90, to help reduce deployment gap. The gripper-mounted camera is located close to the gripper surface, minimizing collisions of the protruding camera when operating in tight space. Gripper openness is measured by the rotary encoders at gripper joints producing sub-millimetre accuracy, without relying on visual identification of gripper openness. Gripper poses are tracked by an external \emph{optical motion-capture system} which resolves each 6-DoF trajectory at sub-millimetre precision in a single global Cartesian frame and also synchronises the head RGB-D camera to avoid interference of IR emissions. This optical tracking replaces the on-board visual SLAM used by conventional UMI rigs, obtaining superior accuracy in pose trajectories with minimal operational risks of pose jitters and track losses due to temporary lack of visual features in SLAM-based pose estimation systems. This setup with optical tracking systems prioritizes high quality action labels for tasks involving fine motor skills, at the cost of inconvenient in-the-wild deployment. The grippers are designed with ergonomic finger-attached mechanisms that allow direct actuation with contact and force feedback onto human fingers rather than relying on indirect trigger-based actuation with less obvious force feedback. Some grippers are optionally instrumented with 6-dimensional force torque sensors located at the tips, and their fingers are \emph{attached to the operator's own fingers} rather than operated through mechanical triggers, giving a proprioception-aligned mapping between human intent and recorded action, and the tip-located sensors allow more direct measurements of force intent compared wrist-located sensors. Because the recording is anchored to the gripper rather than to any fixed base, the corpus is free of base-placement variance.
The rigs capture RGB-D streams, though in the current version of \HyVLA, only the RGB modality is consumed in training, while depth data remain available for future training stages. 

\paragraph{Composition and distribution.} The corpus spans more than $1$M episodes and $10$K hours of demonstrations across 70 distinct tasks, organised into six scene-based task families---Laundry Room (28.5\%), Kitchen (19.2\%), Personal Care \& Miscellaneous (13.8\%), Dexterous / Tool-use (10.4\%), Storage \& Organization (10.0\%), and Cleaning (5.7\%). These six families account for the bulk of the corpus, while the remaining tasks form a long tail spanning diverse object categories and environmental conditions. Manipulated objects cover a broad spectrum from rigid containers and tableware to precision instruments and deformable fabrics. A complete characterization of task families, object-category breakdown, and per-task hour distribution is provided in Fig.~\ref{fig:umi_data_dist}.

\subsection{Pre-training}
\label{sec:pretrain-recipe}

\paragrapha{Setup.} 
We initialize the VLM from the Hy-Embodied-0.5-MoT~\citep{hyembodied2025} checkpoint for pre-training. While the action expert shares the same architectural configuration as the VLM, it is instantiated and randomly initialized as an independent Transformer module. Furthermore, its hidden and intermediate sizes are scaled down from $2048$ to $1024$ and from $6144$ to $2048$, respectively, yielding an effective parameter count of $370$M.
All model parameters are trainable and are optimized under the flow-matching objective (Eq.~\ref{eq:fm}). To accelerate large-scale pre-training, we set $K{=}1$, i.e., no historical image frames are used as input, so the video encoder (Sec.~\ref{sec:short-term-mem}) reduces to the standard single-image encoder. The policy ingests $3$ camera views at $224{\times}320$ resolution and predicts a future action chunk of horizon $H{=}50$ at $10$\,Hz.

\paragrapha{Data and Pre-training recipe.} We use the full $10$K-hour UMI corpus for pre-training. The dataloader samples the dataset with replacement: it first samples an episode from the full corpus with probability proportional to episode length, then uniformly samples one frame from that episode as the current frame, and finally takes the future action sequence with chunk size $H{=}50$ at $10$\,Hz as the ground-truth action chunk. Both state and action inputs are normalized using their dataset-wide mean and standard deviation before being fed into the network. We train for $200$K steps with a global batch size of $1{,}024$ and a base learning rate of $5\times10^{-5}$. The learning rate is linearly warmed up to its maximum value over the first $1$K steps, decayed to one tenth of the peak value over the subsequent $160$K steps, and kept training for another $40$K steps. We use AdamW optimizer~\cite{loshchilov2019decoupledweightdecayregularization} and perform training in \texttt{bfloat16} mixed precision.

\subsection{Supervised Fine-tuning}
\label{sec:stage2}

\paragrapha{Setup.} Initializing from the UMI pre-trained VLA checkpoint (Sec.~\ref{sec:pretrain-recipe}), we run supervised fine-tuning (SFT) on task-specific demonstrations from each target embodiment under the flow-matching objective (Eq.~\ref{eq:fm}). Both the VLM and action expert weights are loaded from the pre-trained model, and all parameters remain trainable. Unlike pre-training, SFT sets $K{=}6$, enabling the video encoder of Sec.~\ref{sec:short-term-mem} to condition on the current frame together with five historical frames.

\paragrapha{Embodiments and Data.} We fine-tune our model across one simulated embodiment and four real-world platforms. In simulation, we employ the Aloha-AgileX bimanual setup from the RoboTwin\,2.0 benchmark~\citep{chen2025robotwin2}, covering $50$ manipulation tasks; each task provides $50$ clean-environment episodes and $500$ randomized-environment episodes, resulting in $2.75$K episodes and more than $6$M frames in total. For real-world SFT, we organize the data into two deployment tracks that separate intra-embodiment adaptation from cross-embodiment transfer. \textbf{Track~A (intra-embodiment)} collects demonstrations through tele-operation on the same robot platform used for evaluation; here, the Dobot X-Trainer covers four tasks with $300$ demonstrations per task ($18$ hours in total). \textbf{Track~B (cross-embodiment)} fine-tunes only on task-specific UMI demonstrations and deploys to morphologically different target robots without target-robot teleoperation; this track covers one task on JAKA\,K1 ($300$ UMI demonstrations, $1.2$ hours) and one task on Astribot\,S1 ($200$ UMI demonstrations, $1.5$ hours). Separately, we use Unitree G1 ($1$ task, $400$ UMI demonstrations, $2.2$ hours) for the force-modality validation in Sec.~\ref{sec:headline_real_world}.

\paragrapha{Post-training recipe.} For \textbf{real-world deployment}, actions are sampled at $50$\,Hz with an action-chunk horizon of $H{=}50$ and a history interval of $1$ second. We train for $60$K steps with a global batch size of $32$ and a base learning rate of $2.5\times10^{-5}$, decayed over $40$K steps. For \textbf{RoboTwin\,2.0}, due to the larger data scale, we downsample future actions from the current frame with stride $3$, use an action-chunk horizon of $H{=}20$ and a history interval of $5{\times}$stride. The global batch size is set to $128$ and the remaining optimization settings follow the pre-training recipe. More details are described in Appendix~\ref{app:pertask}.

\section{Reinforcement Learning Post-Training}
\label{sec:rl_posttraining}

After supervised pre-training and SFT (Sec.~\ref{sec:stage2}), \HyVLA further improves real-robot deployment through failure-driven post-training. This stage follows the \textbf{FlowPRO} recipe~\cite{wu2026flowpro}, using a flow-matching-aware preference-optimization loss (\textbf{RPRO}) together with a teleoperated intervention-and-rollback data pipeline. In this way, \HyVLA converts a small number of real-robot corrections into measurable deployment gains without training any reward or value model.

\subsection{Design Principles}
\label{sec:rl_motivation}

As discussed in section \ref{sec:related}, real-robot post-training generally falls into three families: SFT/DAgger, reward- or value-based RL, and preference-based RL. Their characteristic limitations motivate the three FlowPRO design principles below:
\begin{itemize}[leftmargin=1.6em,itemsep=2pt,topsep=2pt]
  \item \textbf{(P1) Exploit failures directly.} Negative trajectories are not discarded or merely flagged for re-labelling; they are fed back into the action-generation loss as per-state, per-chunk \emph{contrastive} signals against their paired positive corrections.
  \item \textbf{(P2) Avoid reward and critic models entirely.} The training signal is computed in closed form from a frozen reference policy and the current policy, using a flow-matching log-likelihood proxy. This bypasses the dense-reward-design bottleneck that plagues contact-rich manipulation.
  \item \textbf{(P3) Anchor the implicit reward.} A symmetric proximal regularizer prevents the absolute magnitude of the implicit reward from exploding. This structurally forbids the plain-DPO reward-hacking failure mode in which the policy drifts away from \emph{both} $a^w$ and $a^l$.
\end{itemize}
The remainder of this section formalises the loss (Sec.~\ref{sec:rl_method}) and the data pipeline that supplies the per-state preference tuples it consumes.

\subsection{Method}
\label{sec:rl_method}


\begin{figure}[t]
    \centering
    \includegraphics[width=0.92\linewidth]{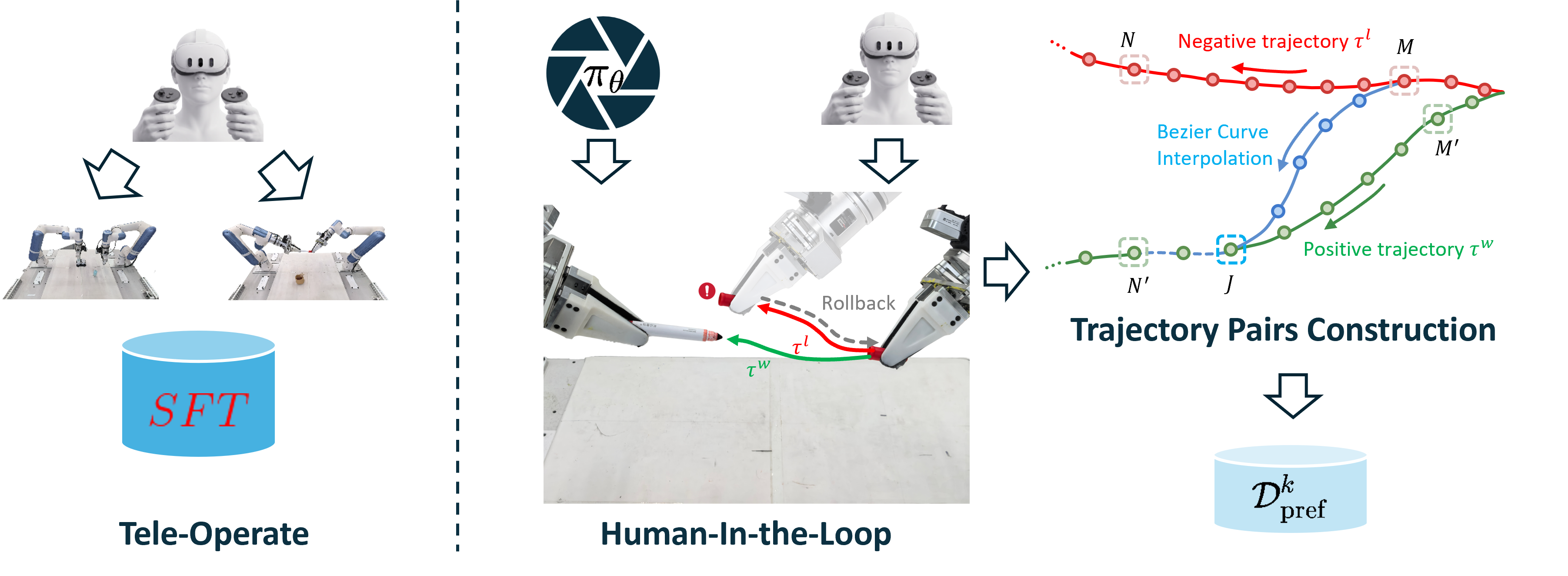}
    \caption{\textbf{FlowPRO data pipeline} for collecting real-robot preference trajectories and converting them into dense per-state preference tuples. During policy rollouts, an operator triggers an intervention-and-rollback: the system rewinds to a prior state, logs the executed segment as a negative trajectory, and records a corrective teleoperation segment as the paired positive trajectory. A smooth-interpolation procedure then synthesizes the missing counterpart action on each branch to yield per-state tuples $(s, a^w, a^l)$ used for preference optimization.}
    \label{fig:rl_train_pipeline}
\end{figure}

\begin{figure}[t]
    \centering
    \includegraphics[width=0.92\linewidth]{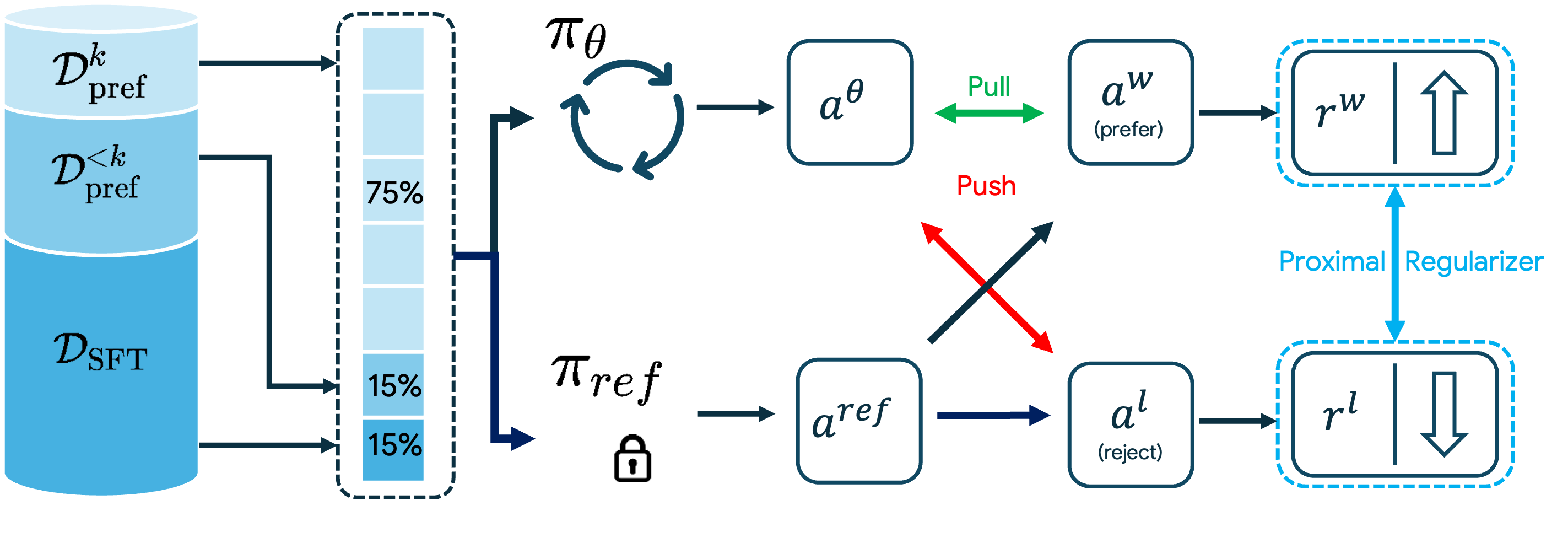}
    \caption{\textbf{RPRO optimization.} The learnable policy $\pi_\theta$ and frozen reference $\pi_{\text{ref}}$ predict actions $a^\theta$ and $a^{\text{ref}}$ for the same state. The objective \emph{pulls} $a^\theta$ toward the preferred action $a^w$ ($r^w\!\uparrow$) and \emph{pushes} it from the dispreferred $a^l$ ($r^l\!\downarrow$). A proximal regularizer (blue dashed) anchors both reward branches to $\pi_{\text{ref}}$, preventing reward hacking. Batches mix $\dataset_{\text{pref}}^{k}$, $\dataset_{\text{pref}}^{<k}$, and $\dataset_{\text{SFT}}$.}
    \label{fig:rl_data_pipeline}
\end{figure}

FlowPRO proceeds as an iterative offline-RL loop on top of an SFT-pretrained \HyVLA base policy (Fig.~\ref{fig:rl_train_pipeline}). Each round contains three steps: (1) collect on-robot preference pairs via teleoperated intervention-and-rollback; (2) convert these sparse trajectory-level corrections into dense per-state preference tuples via Smooth Interpolation; (3) optimize the policy with the RPRO loss on a mixed batch of new pairs, historical pairs, and SFT data (Fig.~\ref{fig:rl_data_pipeline}). The previous round's policy serves as the reference policy $\refpolicy$ in the next round.

\textbf{RPRO loss.}
The \HyVLA action head is a flow-matching model~\citep{lipman2023fm,liu2023flowstraight}. Given a state $s=(o,l)$ with visual observations $o$ and a language instruction $l$, a velocity field $v_\theta(a_t,t\mid s)$ transports Gaussian noise $\epsilon\sim\mathcal{N}(0,I)$ to an action chunk $a$. This transport follows the linear interpolant $a_t=(1-t)\epsilon+ta$ over flow time $t\in[0,1]$, with conditional velocity $u(a_t\mid a):=a-\epsilon$. Following Flow-DPO~\citep{liu2025flowdpo}, we adopt the per-sample flow-matching regression loss as a tractable surrogate for the negative log-likelihood,
\begin{equation}
\ell_\theta(s,a)=\expect_{t\sim\mathcal{U}[0,1],\,\epsilon\sim\mathcal{N}(0,I)}\big[\|v_\theta(a_t,t\mid s)-u(a_t\mid a)\|^2\big],
\label{eq:rl_flow_loss}
\end{equation}
which yields the implicit-reward proxy used by RPRO,
\begin{equation}
r_\theta(s,a)=\tfrac{\beta}{2}\big(\ell_{\text{ref}}(s,a)-\ell_\theta(s,a)\big),
\label{eq:rl_reward_proxy}
\end{equation}
where $\ell_{\text{ref}}$ and $\ell_\theta$ denote the flow-matching losses under the reference and current policies. Substituting Eq.~\eqref{eq:rl_reward_proxy} into the PRO pairwise objective~\citep{guo2025pro} gives the flow-matching-adapted PRO loss,
\begin{align}
\loss_{\text{PRO}}(\theta)=-\expect_{(s,a^w,a^l)\sim\dataset}\Big[
&\underbrace{\log\sigma\!\big(r_\theta(s,a^w)-r_\theta(s,a^l)\big)}_{\loss_{\text{con}}:\ \text{contrastive optimizer}} \nonumber\\
+\;&\underbrace{\sum_{a\in\{a^w,a^l\}}\tfrac{1}{2}\big[\log\sigma\!\big(r_\theta(s,a)\big)+\log\sigma\!\big(-r_\theta(s,a)\big)\big]}_{\loss_{\text{reg}}:\ \text{proximal regularizer}}\Big],
\label{eq:rl_pro_loss}
\end{align}
where $\loss_{\text{reg}}$ is minimized at $r_\theta(s,a)=0$ and grows symmetrically with $|r_\theta(s,a)|$, anchoring the absolute magnitude of the implicit reward and thereby preventing the reward-hacking pathology of plain Flow-DPO. To preserve base-policy performance and reinforce direct regression toward $a^w$, we combine $\loss_{\text{PRO}}$ with a supervised term:
\begin{equation}
\loss_{\text{RPRO}}(\theta)=\lambda_{\text{PRO}}\,\loss_{\text{PRO}}(\theta)+\lambda_{\text{SFT}}\,\loss_{\text{SFT}}(\theta),\qquad \loss_{\text{SFT}}(\theta)=\expect_{(s,a^w)\sim\dataset}[\ell_\theta(s,a^w)].
\label{eq:rl_rpro_loss}
\end{equation}
A useful side-property of Eq.~\eqref{eq:rl_pro_loss} is \emph{contrastive gradient cancellation}: when $a^w=a^l$, $\nabla_\theta\loss_{\text{con}}=\bm{0}$, leaving only $\nabla_\theta\loss_{\text{reg}}$ and $\nabla_\theta\loss_{\text{SFT}}$ active. This makes it safe to route SFT-style demonstrations through the same RPRO loss, which we exploit in batch composition below.

\textbf{Data collection: intervention-and-rollback.}
We collect preference trajectory pairs $(\tau^w,\tau^l)$ via a teleoperated intervention-and-rollback pipeline (Fig.~\ref{fig:rl_train_pipeline}). During rollouts of the current policy, the operator intervenes whenever an erroneous or dangerous action is observed. The system then (1)~rewinds to an earlier state $s_{t-\Delta}$ with operator-chosen horizon $\Delta$ and records the executed segment as the negative trajectory $\tau^l$; (2)~retrieves the observation at $t-\Delta$ as a visual reference in case the environment has changed and the physical scene needs to be reset; and (3)~records the operator's corrective demonstration from $s_{t-\Delta}$ as the positive trajectory $\tau^w$. A single operator action thus yields a naturally paired $(\tau^w,\tau^l)$ sharing the same initial state. Varying $\Delta$ across interventions diversifies the per-pair starting state without recording separate positive and negative rollouts.

\textbf{Smooth Interpolation and batch mixing.}
Because $\tau^w$ and $\tau^l$ diverge after $s_{t-\Delta}$, each subsequent state belongs to only one trajectory. To produce dense per-state tuples $(s,a^w,a^l)$ required by Eq.~\eqref{eq:rl_pro_loss}, we synthesize the missing counterpart with a Smooth Interpolation procedure (Fig.~\ref{fig:rl_train_pipeline}). For a state $M$ on $\tau^l$, we locate its closest point $M'$ on $\tau^w$ under a weighted distance metric. We then construct a synthetic positive action chunk that bridges from $M$ to a transition point $J$ on $\tau^w$ via a cubic B\'ezier for positions, Slerp for orientations, and linear interpolation for the gripper. The chunk then follows $\tau^w$ until it ends at $N'$, while the negative action is simply the next $H$ steps along $\tau^l$. For states already on $\tau^w$ or in $\dataset_{\text{SFT}}$, we set $a^w=a^l$. The contrastive gradient cancellation above makes these samples act as regularized SFT samples. Across iterations, we keep the round-$k$ pairs $\dataset_{\text{pref}}^k$, the historical pool $\dataset_{\text{pref}}^{<k}\!=\!\bigcup_{j<k}\dataset_{\text{pref}}^j$, and $\dataset_{\text{SFT}}$ in separate buffers. We mix mini-batches at fixed proportions: $80\%/20\%$ for $k{=}1$ ($\dataset_{\text{pref}}^k/\dataset_{\text{SFT}}$) and $70\%/15\%/15\%$ for $k{\geq}2$ ($\dataset_{\text{pref}}^k/\dataset_{\text{pref}}^{<k}/\dataset_{\text{SFT}}$). This schedule up-weights the newest, most informative failure states, replays previously corrected ones to prevent regression, and retains a non-trivial SFT share to anchor base capabilities.

\noindent\emph{Experimental validation of FlowPRO on four real-robot bimanual tasks (\textsc{Bottle}, \textsc{Cap}, \textsc{USB}, \textsc{Zip}) is reported in \ref{sec:rl_experiments} together with the rest of the empirical evaluation.}


\section{Deployment}
\label{sec:inference}

Deployment mainly addresses three runtime issues: mapping end-effector delta chunks to heterogeneous robot platforms, serving VLA predictions at the robot control rate, and stitching independently predicted chunks into smooth motion. We handle them with three lightweight components: a platform mapper that keeps the learned action interface unchanged across embodiments (Sec.~\ref{sec:deployment}); an asynchronous inference--execution loop that overlaps backbone forward passes with servo execution (Sec.~\ref{sec:inf_async}); and a latency-aware cubic-B\'{e}zier stitcher that removes stale prefixes and enforces smooth chunk transitions (Sec.~\ref{sec:inf_bezier}). The same deployment stack is used across all real-robot evaluations.

\subsection{Embodiment-Agnostic Platform Mapping}
\label{sec:robots}
\label{sec:deployment}

The role of platform mapping is to preserve the robot-agnostic contract established by the delta-chunk representation. The policy outputs a 20-dimensional dual-arm action chunk (10 dimensions per end-effector: a 3-D Cartesian translation and a 6-D rotation — the first two rows of an $SO(3)$ rotation matrix — both expressed relative to the end-effector pose at the start of the chunk, together with a 1-D gripper opening command). Embodiment-specific kinematics are deferred to deployment, where the relative $SE(3)$ prediction is composed with the initial end-effector pose to recover absolute world-frame targets and inverse kinematics (IK) is then solved on the target robot to produce joint commands.

For intra-embodiment deployment (Track~A), the world frame remains the same in deployment and data collection. For cross-embodiment deployment (Track~B), data collection and deployment use different embodiments, so we instantiate mappings for two embodiment types: fixed-base arms and the floating-base humanoid. In the equations below, ${}^AT_B$ denotes the pose of frame $B$ in frame $A$; $W$, $G_t$, $G_{t+k}$, and $C$ are the world, current gripper, predicted future gripper k steps further, and the chassis frame.

For fixed-base arms such as JAKA\,K1, the rel-EE chunk is cast into the world frame using the current gripper pose ${}^WT_{G_t}$ from forward kinematics,
\begin{equation}
  {}^WT_{G_{t+k}} \;=\; {}^WT_{G_t} \cdot {}^{G_t}T_{G_{t+k}},
  \label{eq:T1}
\end{equation}
For the humanoid like Astribot\,S1, a deterministic heuristic infers a fixed chassis frame ${}^WT_{C}$ and a floating torso frame from the predicted gripper targets (Appendix~\ref{app:deployment}), yielding
\begin{equation}
  {}^{C}T_{G_{t+k}} \;=\; \bigl({}^WT_{C}\bigr)^{-1} \cdot {}^WT_{G_t} \cdot {}^{G_t}T_{G_{t+k}}.
  \label{eq:humanoid}
\end{equation}
where ${}^WT_{C}$ is a constant transform cached after calculation. The additional $24$ head/torso dimensions ($12$ each: $3$ position $+$ $9$ rotation) are set by this heuristic rather than predicted by the policy. This keeps the learned action interface unchanged across Track~A intra-embodiment deployment and Track~B cross-embodiment deployment.

\subsection{Asynchronous Execution for Real-Time Control}
\label{sec:inf_async}

\begin{figure}[t]
    \centering
    \includegraphics[width=\linewidth]{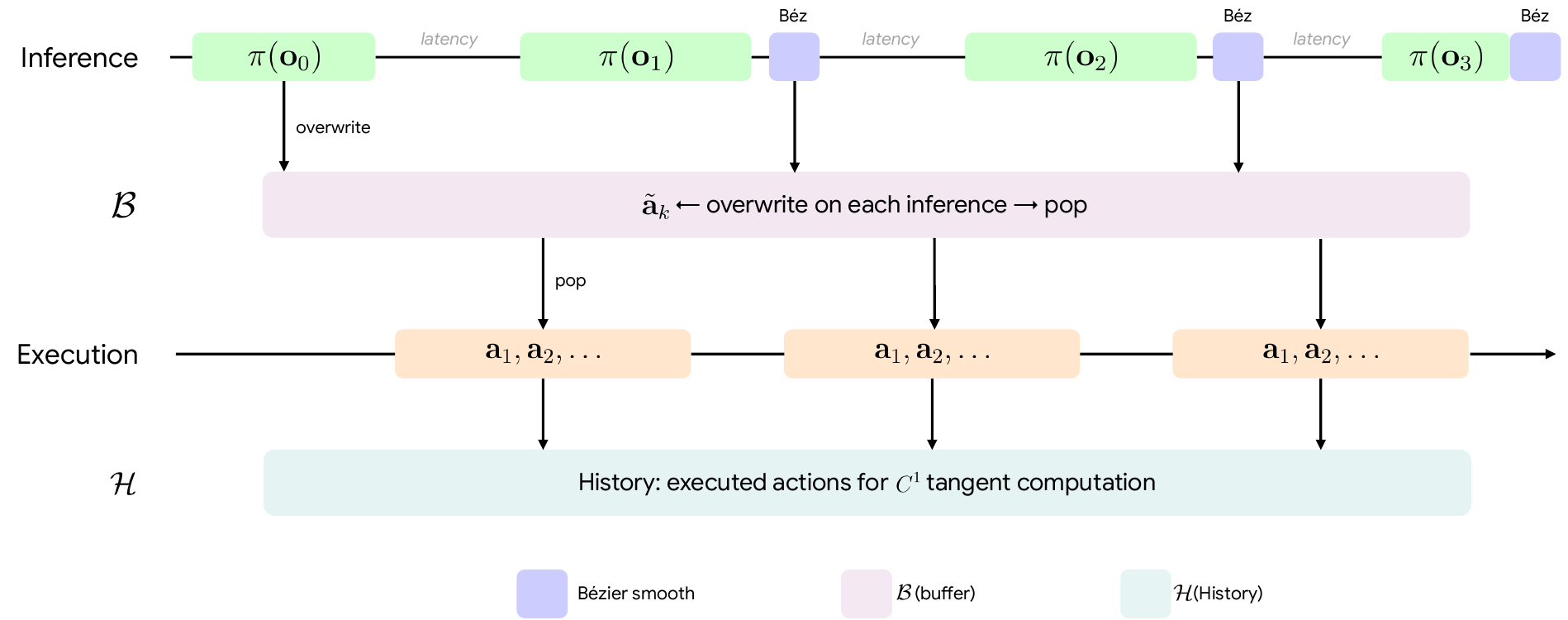}
    \caption{\textbf{Asynchronous execution timeline.} Policy inference, B\'{e}zier smoothing, buffer overwrite, and servo-rate action execution are overlapped; executed actions are recorded in $\mathcal{H}$ to estimate tangents for the next chunk stitch.}
    \label{fig:async_pipeline}
\end{figure}

A high-capacity VLA policy runs slower than the robot servo loop, so synchronous execution would leave the robot idle between forward passes. We therefore decouple inference from command dispatch using a producer--consumer runtime with a thread-safe action buffer $\mathcal{B}$ (Fig.~\ref{fig:async_pipeline}). The inference thread queries the policy from the latest observation and overwrites $\mathcal{B}$ with a smoothed action sequence, while the execution thread pops commands from $\mathcal{B}$ at the control frequency and records recent poses for tangent estimation. Overlapping these two loops hides much of the backbone latency behind continuous execution.

\subsection{Latency-Aware B\'{e}zier Chunk Stitching}
\label{sec:inf_bezier}

Chunk stitching is critical in asynchronous execution: delayed chunks must be reconnected to the robot's current state without introducing motion discontinuities. We use a cubic B\'{e}zier segment to form a compact $C^1$-continuous connector with controllable endpoint positions and tangents.

The first design choice is to select the connection point between the B\'{e}zier connector and the retained chunk. We set $\gamma$ as a lightweight deployment hyperparameter, chosen according to hardware response such as acceleration limits and servo rate. A smaller $\gamma$ selects an earlier point in the retained chunk and preserves more policy-predicted actions, but leaves less room to correct the delayed boundary. A larger $\gamma$ provides a smoother landing target but skips more predicted actions. We clip the resulting index away from both ends so that the future tangent can be estimated from neighboring waypoints.

The control points are chosen with the same intuition. The B\'{e}zier curve should leave the robot's current trajectory in the direction it was already moving, and enter the future chunk in the direction that the policy predicts next. We therefore place the two inner control points along the historical motion direction and the local direction of the future chunk, with their distance scaled by the gap between the current robot state and the reconnection point. This gives a smooth transition without introducing an additional learned controller.

Based on this design, the runtime procedure is as follows. Given an original chunk of length $N$, we first discard the stale prefix
\begin{equation}
    K = \lceil N/\alpha \rceil, \qquad K \leq N-3,
    \label{eq:truncation}
\end{equation}
where $\alpha>1$ is the truncation ratio, and retain $\mathcal{F}=\{\mathbf{f}_0,\ldots,\mathbf{f}_{M-1}\}$ with $M=N-K$.

Let $\mathbf{h}_0$ be the last executed EE position. We choose an interior connection point $\mathbf{f}_c$ with $c=\mathrm{clip}(\lfloor\gamma M\rfloor,1,M-2)$, where the clipping keeps the two-sided future tangent well-defined, and construct a cubic B\'{e}zier segment $\mathbf{B}(t)$ satisfying
\begin{align}
    \mathbf{B}(0)=\mathbf{h}_0, \quad
    \mathbf{B}(1)=\mathbf{f}_c, \quad
    \dot{\mathbf{B}}(0) \parallel \hat{\mathbf{d}}_{\mathrm{hist}}, \quad
    \dot{\mathbf{B}}(1) \parallel \hat{\mathbf{d}}_{\mathrm{fut}},
\end{align}
with tangents
\begin{equation}
    \hat{\mathbf{d}}_{\mathrm{hist}} = \frac{\mathbf{h}_0-\mathbf{h}_{-1}}{\|\mathbf{h}_0-\mathbf{h}_{-1}\|}, \qquad
    \hat{\mathbf{d}}_{\mathrm{fut}} = \frac{\mathbf{f}_{c+1}-\mathbf{f}_{c-1}}{\|\mathbf{f}_{c+1}-\mathbf{f}_{c-1}\|},
\end{equation}
when the corresponding norm is non-zero. The endpoint control points anchor the transition at the current state and the reconnection point, while the two inner control points encode the historical and future tangents:
\begin{align}
    \mathbf{P}_0 &= \mathbf{h}_0, &
    \mathbf{P}_1 &= \mathbf{P}_0 + \lambda\hat{\mathbf{d}}_{\mathrm{hist}}, \\
    \mathbf{P}_2 &= \mathbf{P}_3 - \lambda\hat{\mathbf{d}}_{\mathrm{fut}}, &
    \mathbf{P}_3 &= \mathbf{f}_c,
\end{align}
where $\lambda=\sigma\|\mathbf{P}_3-\mathbf{P}_0\|$ controls the tangent length. The transition curve is
\begin{equation}
    \mathbf{B}(t)=(1-t)^3\mathbf{P}_0+3(1-t)^2t\mathbf{P}_1+3(1-t)t^2\mathbf{P}_2+t^3\mathbf{P}_3,
\end{equation}
and is uniformly sampled to replace the discontinuous boundary segment. Position is smoothed in $\mathbb{R}^3$, orientation uses SLERP, gripper commands are linearly interpolated, and each arm is processed independently. The resulting transition is $C^1$-continuous, policy-agnostic, and controlled by the embodiment-dependent parameters $\alpha$, $\gamma$, and $\sigma$.

\begin{figure}[t]
    \centering
    \includegraphics[width=\textwidth]{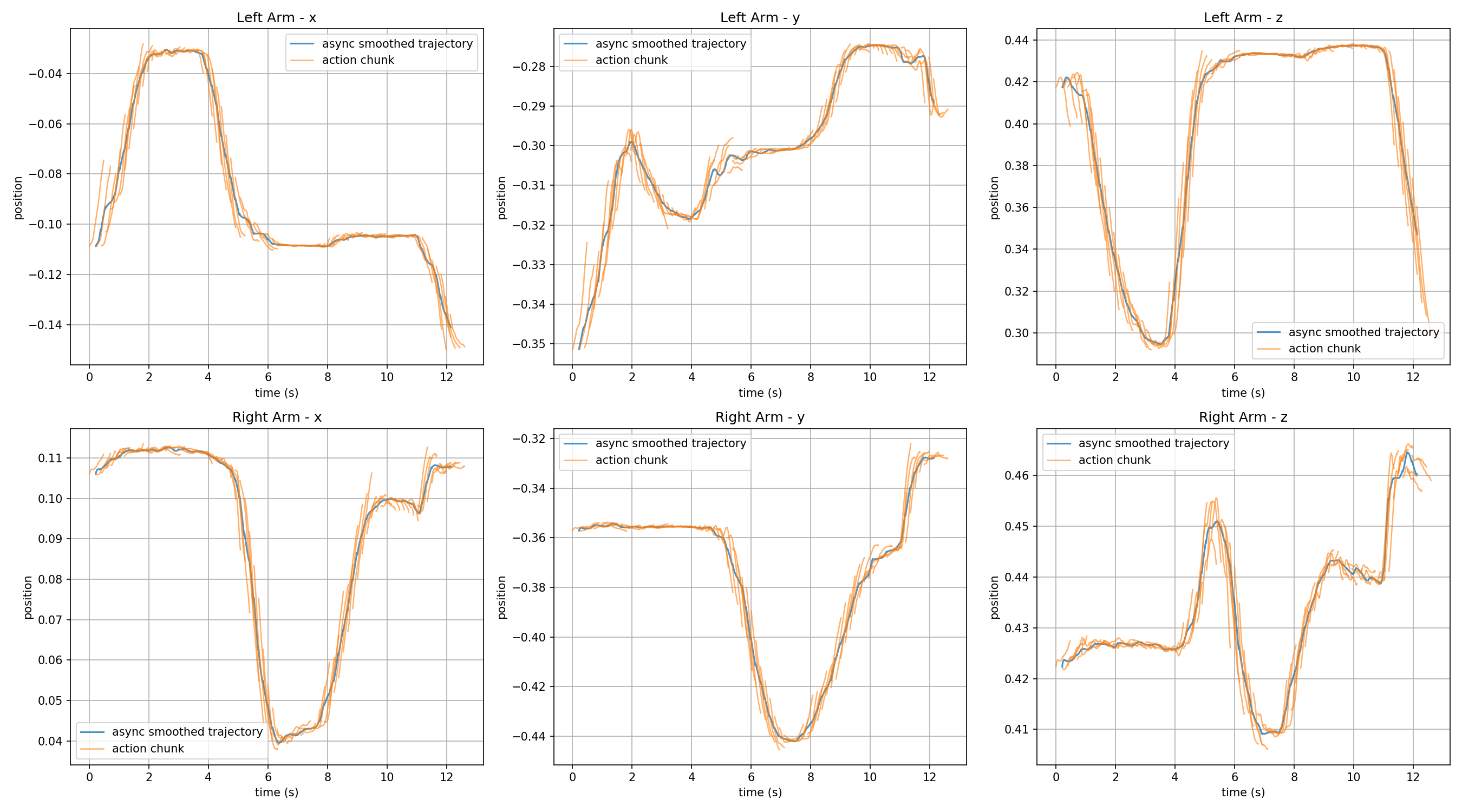}
    \caption{\textbf{Trajectory comparison} between raw action chunks (orange) and the asynchronous B\'{e}zier-smoothed trajectory (blue). Smoothing reduces visible discontinuities at chunk boundaries for both arms across $x$, $y$, and $z$ dimensions.}
    \label{fig:action_comparison}
\end{figure}

\FloatBarrier
Fig.~\ref{fig:action_comparison} compares raw chunked actions with the asynchronously B\'{e}zier-smoothed trajectory.

\section{Evaluation}
\label{sec:experiments}

\label{sec:setup}

The empirical validation addresses two parallel questions: how well \HyVLA performs after standard downstream supervised fine-tuning in simulation and on real hardware (Secs.~\ref{sec:headline_robotwin} and \ref{sec:headline_real_world}), and how much FlowPRO post-training further improves a deployed policy (Sec.~\ref{sec:rl_experiments}).

For real hardware, we organize the SFT evaluation into two deployment tracks: \textbf{Track A} fine-tunes and evaluates on the same tele-operated robot platform, while \textbf{Track B} fine-tunes only on UMI demonstrations and deploys on morphologically different robots without target-robot teleoperation. We evaluate four Track-A tasks and two Track-B tasks. Foundational baselines $\pi_0$ and $\pi_{0.5}$ are identically parameterized and trained with matched data and iteration budgets.

\subsection{Simulated Tasks}
\label{sec:headline_robotwin}

On RoboTwin\,2.0~\citep{chen2025robotwin2}, we report task success rates averaged over $100$ stochastic rollouts per task and then over the full $50$-task suite. Results are evaluated under both \emph{Clean} and \emph{Randomized} settings, with aggregate comparisons and ablations shown in Table~\ref{tab:robotwin_comp}; the complete per-task breakdown is deferred to Appendix~\ref{app:pertask} (Table~\ref{tab:robotwin_pertask}).

\begin{table}[!htbp]
  \centering
  \footnotesize
  \setlength{\tabcolsep}{8pt}
  \renewcommand{\arraystretch}{1.1}
  \begin{tabular}{l c c}
    \toprule
    \textbf{Method} & \textbf{Clean} & \textbf{Randomized} \\
    \midrule
    \multicolumn{3}{@{}l}{\emph{Other methods}} \\
    $\pi_{0}$~\citep{black2024pi0}                     & 65.9 & 58.4 \\
    ABot-M0~\citep{abot_m0_2025}                       & 81.2 & 80.4 \\
    $\pi_{0.5}$~\citep{intelligence2025pi05}           & 82.7 & 76.8 \\
    Qwen-VLA~\citep{qwenvla}                           & 86.1 & 87.2 \\
    LingBot-VLA~\citep{lingbot_vla_2025}               & 86.5 & 85.3 \\
    starVLA~\citep{star_vla_2025}                      & 88.2 & 88.3 \\
    Motus~\citep{motus_2025}                           & 88.7 & 87.0 \\
    JoyAI-RA~\citep{joyai_ra_2025}                     & 90.5 & 89.3 \\
    \midrule
    \multicolumn{3}{@{}l}{\emph{Ablations}} \\
    \textbf{\HyVLA}                  & \textbf{90.9} & \textbf{90.1} \\
    \quad w/o compact memory encoder                                  & 88.8 & 88.6 \\
    \quad w/o compact memory encoder and UMI pre-training              & 88.1 & 87.9 \\
    \bottomrule
  \end{tabular}

  \vspace{4pt}
  \caption{\textbf{Evaluation results on the RoboTwin\,2.0 benchmark~\citep{chen2025robotwin2}.} Success rate (\%) under the Clean and Randomized settings, averaged over $100$ runs per task and then over the $50$-task suite. The upper block lists competing methods; the lower block reports removal-based ablations from the full \textbf{\HyVLA} model. Per column, the best result is in \textbf{bold}.}
  \label{tab:robotwin_comp}
\end{table}

\paragrapha{Baselines and main results. }
We benchmark against eight contemporary VLA systems---$\pi_0$~\citep{black2024pi0}, $\pi_{0.5}$~\citep{intelligence2025pi05}, ABot-M0~\citep{abot_m0_2025}, LingBot-VLA~\citep{lingbot_vla_2025}, starVLA~\citep{star_vla_2025}, Motus~\citep{motus_2025}, JoyAI-RA~\citep{joyai_ra_2025}, and Qwen-VLA~\citep{qwenvla}---using each method's officially reported success rates under the same Clean and Randomized protocol. As shown in the upper block of Table~\ref{tab:robotwin_comp}, \HyVLA attains the best success rate in \emph{both} settings, reaching $90.9\%$ on Clean and $90.1\%$ on Randomized. It outperforms $\pi_0$ by $25.0$ points on Clean (vs.\ $65.9\%$) and by $31.7$ points on Randomized (vs.\ $58.4\%$), and remains clearly ahead of $\pi_{0.5}$ by $8.2$ and $13.3$ points (vs.\ $82.7\%$ and $76.8\%$). Even against the strongest competing method, JoyAI-RA, \HyVLA still leads by $0.4$ and $0.8$ points (vs.\ $90.5\%$ and $89.3\%$).

\paragrapha{Ablation. }
The lower block of Table~\ref{tab:robotwin_comp} conducts a removal-based ablation starting from the full \HyVLA model. Removing the compact memory encoder reduces performance from $90.9\%$\,/\,$90.1\%$ to $88.8\%$\,/\,$88.6\%$ on Clean\,/\,Randomized; further removing the large-scale UMI pre-training stage lowers the scores to $88.1\%$\,/\,$87.9\%$. Together, these ablations show that both UMI pre-training and short-horizon visual memory contribute consistent gains. Although the egocentric real-world UMI corpus is visually distant from the synthetic RoboTwin\,2.0 renderings, UMI pre-training still provides a modest gain in simulation. The limited magnitude is expected, given the large gaps in task distribution, action trajectories, and visual appearance. In contrast, Sec.~\ref{sec:headline_real_world} shows its significant benefit on real-robot tasks, where the domain gap to UMI demonstrations is smaller.

\FloatBarrier

\subsection{Real-World Tasks}
\label{sec:headline_real_world}

\begin{figure}[!htbp]
    \centering
    \includegraphics[width=\textwidth]{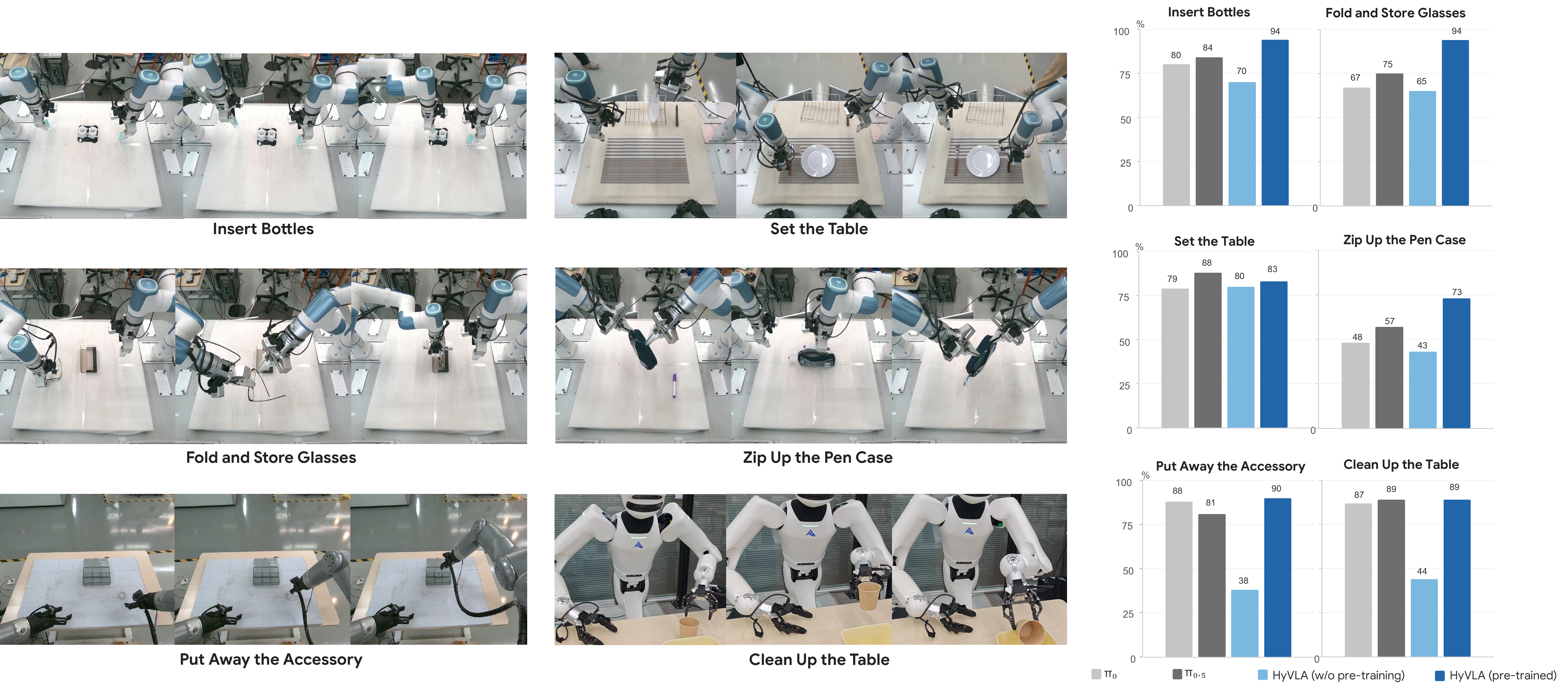}
    \caption{\textbf{Real-robot evaluation on six bimanual manipulation tasks.}
    Left panel: Snapshots of representative task executions captured during rollout.
    Right panel: Per-task success rates (\%) after supervised fine-tuning on tele-operated or UMI demonstrations.}
    \label{fig:real_world_results}
\end{figure}

We evaluate \HyVLA on real-robot bimanual manipulation through the two deployment tracks introduced above, spanning three platforms and six benchmark tasks, plus a qualitative force-discrimination task on a Unitree G1. All per-task snapshots and success rates are reported in Figure~\ref{fig:real_world_results}. Track A tests intra-embodiment fine-tuning, while Track B probes whether UMI-only post-training can transfer task semantics across embodiments.

\paragrapha{Track A --- Intra-Embodiment Fine-Tuning (Dobot X-Trainer).}
Data are collected via tele-operation on a Dobot X-Trainer and the same platform is used for evaluation. We benchmark four bimanual tasks: \textit{Insert Bottles}, where the robot grasps two cylindrical bottles and inserts each into a dedicated holder under tight geometric tolerances; \textit{Fold and Store Glasses}, where it picks up a pair of eyeglasses, folds the temples inward through coordinated bimanual motion, and places them into a protective case; \textit{Set the Table}, where it arranges a plate, a fork, and a knife at canonical positions on a dining surface, requiring long-horizon spatial planning and precise 6-DoF placement; and \textit{Zip Up the Pen Case}, where it opens the zipper, inserts a pen, and closes the zipper along the full track under deformable-object dynamics.

\paragrapha{Effect of UMI Pre-training on Track-A Tasks.}
The per-task results in Figure~\ref{fig:real_world_results} reveal a consistent pattern on the precision-critical tasks. For \emph{Fold and Store Glasses} and \emph{Zip Up the Pen Case}, success hinges on a few decisive sub-steps rather than on the trajectory as a whole, \eg folding the temples without slipping, or pinching the zipper slider before pulling. Without Hy-UMI-10K pre-training, the policy is visibly less accurate at exactly these moments, where sub-centimetre positioning and stable bimanual force coupling are required. The resulting local errors then propagate downstream and dominate the failure modes. Pre-training reverses this pattern: predictions sharpen at the same critical moments, end-to-end success rates rise accordingly, and coarser segments of the trajectory remain essentially unchanged. This task-level evidence corroborates the simulation ablation in Section~\ref{sec:headline_robotwin}. It suggests that the principal value of large-scale, high-precision UMI pre-training is to sharpen the action distribution at the precision-critical bottlenecks of downstream manipulation, and that this benefit transfers from human demonstrations to real-robot post-training.

\paragrapha{Track B --- Cross-Embodiment Transfer (JAKA K1, Astribot S1).}
For each target robot, we post-train the UMI pre-trained checkpoint on task-specific UMI demonstrations only---without any target-robot teleoperation---and deploy the resulting policy on the corresponding robot. We benchmark two tasks: \textit{Put Away the Accessory} on JAKA K1, where the robot picks up a sub-centimetre hair tie and places it into the centre cell of a compartment box whose cell size nearly matches the tie's diameter; and \textit{Clean Up the Table} on Astribot S1, where the humanoid locates scattered paper cups on a tabletop and deposits them sequentially into a waste bin.

\paragrapha{Effect of UMI Pre-training on Track-B Tasks.}
Track B isolates the contribution of UMI pre-training to cross-embodiment deployment: since no target-robot data is ever seen during fine-tuning, any gain over an identically configured baseline must come from the prior learned during pre-training. Figure~\ref{fig:real_world_results} shows that this gain is substantial on both robots: \HyVLA achieves markedly higher success rates than $\pi_0$ and $\pi_{0.5}$ on \textit{Put Away the Accessory} and \textit{Clean Up the Table}, despite all three policies being post-trained on the same UMI data. The improvement indicates that large-scale, high-fidelity UMI pre-training equips the model with embodiment-agnostic action priors that survive a deployment shift to morphologically unseen robots, and that these priors make the small UMI fine-tuning set sufficient on its own to recover deployable performance on a new platform.

\begin{figure}[!htbp]
    \centering
    \includegraphics[width=0.90\textwidth]{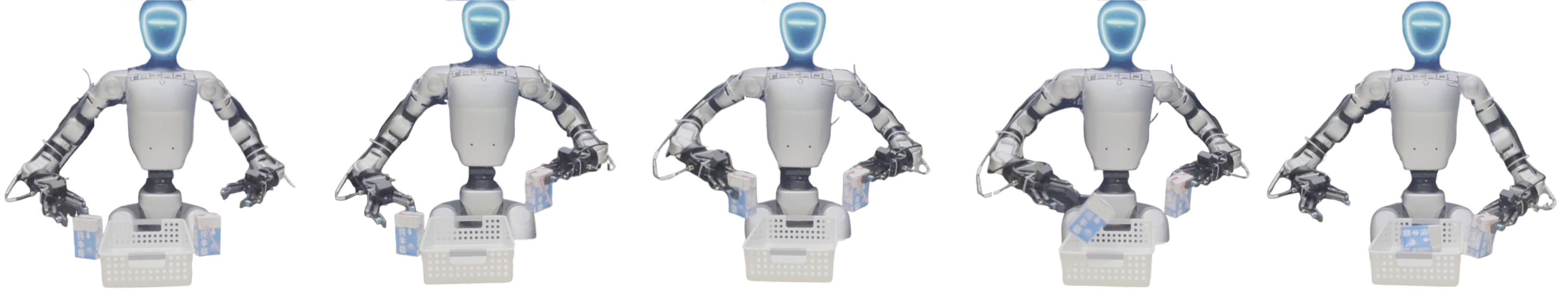}
    \caption{\textbf{Force-guided object discrimination on a Unitree G1.}
    The robot sequentially grasps two boxes of differing mass and places the lighter one into the front basket, confirming that the in-house UMI workstation captures actionable tactile information.}
    \label{fig:force_discrimination}
\end{figure}

\paragrapha{Force-Modality Validation (Unitree G1).}
Because our handheld UMI gripper records tip force signals during demonstration collection, the resulting data directly contains the physical cues needed for force-aware, and potentially force-controlled, manipulation. We show this capability on a Unitree G1 equipped with our end effector, where the policy performs a force-discrimination task: it sequentially grasps two boxes and places the lighter one into a front basket. For this task, we augment the action expert with two lightweight TCN encoders~\cite{lea2017tcn} and an MLP projector, which together encode a $50$-step F/T window for each hand ($\sim$2M added parameters). The augmented policy is then post-trained on a small set of UMI demonstrations recorded with the workstation's tip force/torque signals (Sec.~\ref{sec:umi-corpus}). Since the lighter-object position is randomized across trials, spatial memory alone cannot solve the task; the policy must compare the grasp-phase force profiles before deciding which box to place. \HyVLA reliably selects the lighter box across trials (Fig.~\ref{fig:force_discrimination}), showing that the tactile signals captured by the UMI workstation provide actionable non-visual cues for downstream policy learning.

\subsection{Real-World Reinforcement}
\label{sec:rl_experiments}

\begin{figure}[!htbp]
    \centering
    \includegraphics[width=\textwidth]{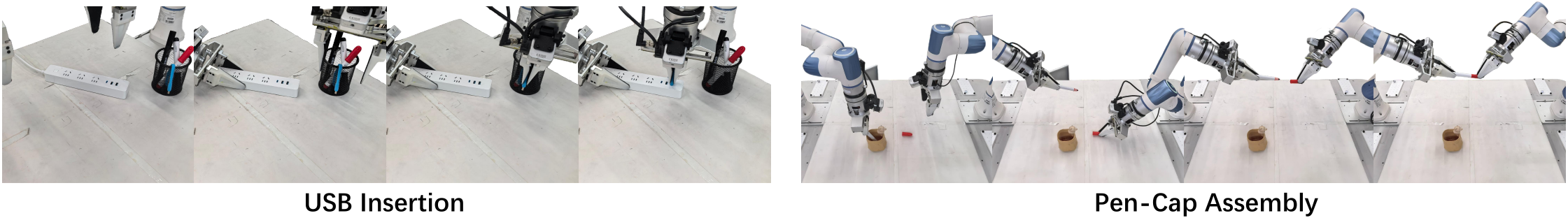}
    \caption{\textbf{Additional fine-grained real-robot tasks for FlowPRO post-training.} Beyond \textbf{Insert Bottles} (\textsc{Bottle}, sub-cm insertion) and \textbf{Zip Up the Pen Case}, which are shown in Fig.~\ref{fig:real_world_results}, we further evaluate FlowPRO on two fine-grained tasks: \textbf{USB insertion} (\textsc{USB}, sub-mm precision) and \textbf{Pen-Cap Assembly} (\textsc{Cap}, in-air bimanual coordination). This figure illustrates these two additional tasks.}
    \label{fig:rl_tasks}
\end{figure}

\textbf{Setup.}
All FlowPRO experiments are conducted on a Dobot X-Trainer bimanual platform. We evaluate on four long-horizon bimanual tasks (Fig.~\ref{fig:rl_tasks}): \textsc{Bottle}, \textsc{Cap}, \textsc{USB}, and \textsc{Zip}. Starting from the same \HyVLA SFT checkpoint $\refpolicy$, every method runs $K{=}3$ rounds of iterative post-training under an identical data-collection budget. Each entry in Table~\ref{tab:rl_main} is averaged over $3$ training seeds; per-seed success rate (SR) is computed from $n{=}100$ rollouts with randomized initial placements, and completion time (CT) is averaged over the same rollouts.

\textbf{Baselines.}
We compare RPRO against two representative comparators that cover both regimes of the design space: \textbf{DAgger}~\citep{ross2011dagger} (positive-only dataset aggregation) and \textbf{$\pi_{0.6}$*}~\citep{intelligence2025pi06} (advantage-conditioned regression that uses the same positive-and-negative pairs as RPRO but injects the preference signal as a conditioning token rather than through a contrastive loss). All methods share the same \HyVLA SFT backbone and the same iterative data-collection protocol.

\begin{table}[!htbp]
\centering
\setlength{\tabcolsep}{4pt}
\begin{tabular*}{\textwidth}{@{\extracolsep{\fill}}lcccccccc@{}}
\toprule
\multirow{2}{*}{Fine-tune} & \multicolumn{2}{c}{\textsc{Bottle}} & \multicolumn{2}{c}{\textsc{Cap}} & \multicolumn{2}{c}{\textsc{USB}} & \multicolumn{2}{c}{\textsc{Zip}} \\
\cmidrule(lr){2-3}\cmidrule(lr){4-5}\cmidrule(lr){6-7}\cmidrule(lr){8-9}
                          & SR & CT & SR & CT & SR & CT & SR & CT \\
\midrule
DAgger                    & $\,$93$\,\pm\,$2.1$\,$\% & 27\,s & $\,$88$\,\pm\,$1.8$\,$\% & 29\,s & $\,$86$\,\pm\,$2.4$\,$\% & 25\,s & $\,$83$\,\pm\,$2.0$\,$\% & 55\,s \\
$\pi_{0.6}$*              & $\,$95$\,\pm\,$1.5$\,$\% & 24\,s & $\,$95$\,\pm\,$1.2$\,$\% & 27\,s & $\,$95$\,\pm\,$1.4$\,$\% & 23\,s & $\,$89$\,\pm\,$1.6$\,$\% & 45\,s \\
\textbf{RPRO}             & $\bm{99\!\pm\!0.6}$\textbf{\%} & \textbf{16\,s} & $\bm{99\!\pm\!0.7}$\textbf{\%} & \textbf{21\,s} & $\bm{98\!\pm\!0.9}$\textbf{\%} & \textbf{22\,s} & $\bm{94\!\pm\!1.1}$\textbf{\%} & \textbf{37\,s} \\
\bottomrule
\end{tabular*}
\vspace{4pt}
\caption{\textbf{Final success rate and completion time after $K{=}3$ rounds of post-training on four real-robot bimanual tasks, with \HyVLA as the base policy.} SR ($\uparrow$, \%) is reported as mean~$\pm$~std (in points) across $3$ training seeds, with each per-seed SR computed over $n{=}100$ randomized rollouts; CT ($\downarrow$, s) is the cross-rollout mean. Best per column in \textbf{bold}.}
\label{tab:rl_main}
\end{table}

\begin{figure}[!htbp]
    \centering
    \includegraphics[width=\textwidth]{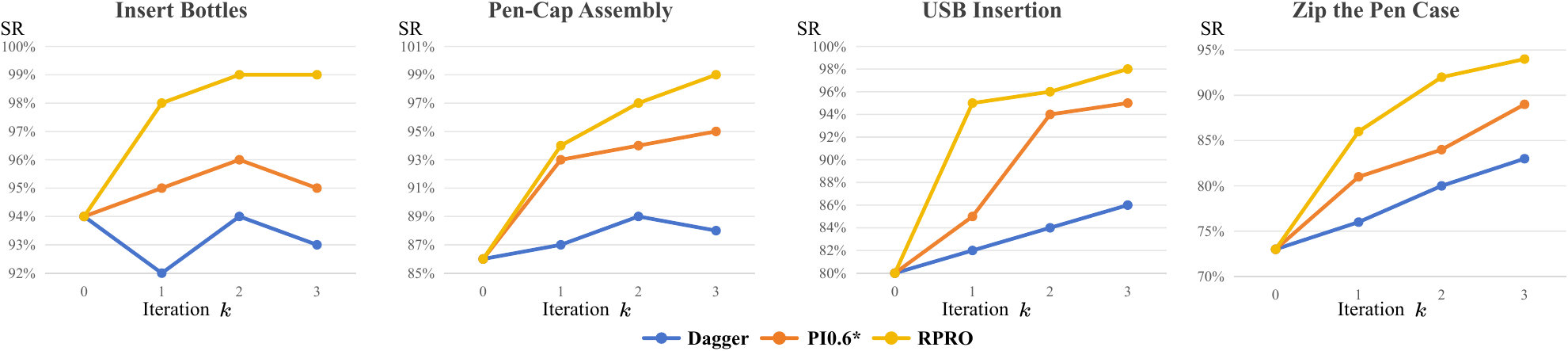}
    \caption{\textbf{Per-iteration success rate on the four real-robot tasks with \HyVLA as the base policy.} Iteration~$0$ corresponds to the shared SFT checkpoint; iterations $1$--$3$ correspond to successive rounds of post-training. RPRO consistently dominates DAgger and $\pi_{0.6}$* throughout the iterative process.}
    \label{fig:rl_hy_compare}
\end{figure}

\textbf{Results.}
Table~\ref{tab:rl_main} and Fig.~\ref{fig:rl_hy_compare} summarize the comparison.
\emph{RPRO vs.\ DAgger.} DAgger relies on positive samples only, while RPRO additionally exploits negative trajectories through a contrastive loss; the resulting per-state push-away gradient from $a^l$ pulls the policy back from nearby failure modes, yielding a consistent gain across all four tasks.
\emph{RPRO vs.\ $\pi_{0.6}$*.} On \emph{identical} preference data, RPRO still outperforms the advantage-conditioned $\pi_{0.6}$* baseline. $\pi_{0.6}$* relies on the model to discover the ``improved''/``unimproved'' partition from a single conditioning token under a pure regression objective---an indirect pressure that can be diluted by the rest of the VLM context---whereas RPRO (\ref{sec:rl_method}) injects the preference signal directly into the action-generation loss, pushing $\policy$ toward $a^w$ and away from $a^l$ per state and per chunk. Across all four tasks, RPRO attains the highest SR with the shortest CT, indicating both more reliable and more efficient task execution.

\FloatBarrier

\section{Related Work}
\label{sec:related}

\paragraph{Generalist VLA Models}
Early VLAs abstracted robotic control into discrete tokens processed by autoregressive heads atop pre-trained VLMs, as exemplified by RT-2~\citep{brohan2023rt2} and OpenVLA~\citep{kim2024openvla}. While effectively transferring semantic priors, this discretisation inherently constrained control frequency and spatial precision. $\pi_0$~\citep{black2024pi0} supplanted discrete action spaces with flow-matching velocity fields, restoring continuous, high-frequency (e.g., $50$\,Hz) execution capabilities. Concurrently, DeepMind introduced Gemini Robotics~\citep{team2025gemini}, bringing Gemini-level reasoning to physical control, and NVIDIA released GR00T N1~\citep{bjorck2025gr00t}, an open foundation model for generalist humanoid control pre-trained on teleoperation, human video, and synthetic data. $\pi_{0.5}$~\citep{intelligence2025pi05} subsequently advanced the flow-matching paradigm with open-world generalization, while Gemini Robotics 1.5~\citep{team2025gemini15} extended the approach with advanced embodied reasoning and cross-embodiment motion transfer. LingBot-VLA~\citep{lingbot_vla_2025} takes a pragmatic approach, scaling to 20\,K hours of real-world dual-arm data across 100 tasks with a throughput-optimised open-source codebase. Unlike autoregressive VLAs, \HyVLA operates entirely within a continuous flow-matching paradigm; unlike $\pi_0$ and $\pi_{0.5}$, it adopts an MoT-based embodied-native backbone, relies on a 10\,K-hour UMI pre-training corpus, and features a specialised deployment protocol for zero-shot cross-embodiment transfer.

\paragraph{Embodied VLM Backbones}
The majority of contemporary VLAs depend on general-purpose vision-language models such as PaliGemma~\citep{beyer2024paligemma} or Qwen-VL~\citep{bai2025qwen3vl}. Recently, domain-specific backbones such as RoboBrain~\citep{tan2026robobrain}, RynnBrain~\citep{dang2026rynnbrain}, and Hy-Embodied-0.5~\citep{hyembodied2025} have emerged to better address the fine-grained visual acuity required for manipulation. The internal Hy-Embodied report introduced a prototype VLA fine-tuned from $5$\,k hours of UMI data, achieving promising baseline success rates on X-Trainer tasks~\citep{hyembodied2025}. Building strictly upon this foundation, \HyVLA doubles the UMI scale to $10$\,k hours, implements the rel-EE representation to facilitate humanoid deployment, and introduces the FlowPRO RL post-training stage, thereby extending efficacy to unseen platforms including JAKA and Astribot\,S1.

\paragraph{Pre-training and Post-training Recipes for VLAs}
Current multi-embodiment pre-training paradigms, such as TRI's Large Behaviour Models~\citep{tri2024lbm} and the $\pi_{0.5}$ methodology, primarily leverage aggregated teleoperation datasets (e.g., Open-X-Embodiment~\citep{collaboration2024oxe}, DROID~\citep{khazatsky2024droid}). By contrast, the foundational pre-training signal of \HyVLA is predominantly sourced from human-centric UMI data, optimising the action expert under a singular flow-matching loss.

\paragraph{Hand-Held Demonstrations and UMI}
The Universal Manipulation Interface (UMI)~\citep{chi2024umi} pioneered the capture of robot-agnostic demonstration data via hand-held gripper rigs. Subsequent efforts, such as DexUMI~\citep{xu2025dexumi}, expanded the morphological applicability of such rigs. Recently, several frameworks have addressed the feasibility of migrating UMI-style hand-held data to humanoid and mobile systems, such as EgoMI~\citep{yu2025egomi} (which captures synchronized head-hand tracking for whole-body and active vision manipulation) and HoMMI~\citep{xu2026hommi} (which learns whole-body mobile manipulation directly from robot-free egocentric human demonstrations). \HyVLA scales UMI data to over 10k hours, and demonstrates UMI-based cross-embodiment transfer to a humanoid under a Stage-2 protocol entirely devoid of target-robot teleoperation.

\paragraph{Preference Post-Training in Continuous Control}
Real-robot post-training pipelines for VLA models broadly fall into three families, each with a characteristic limitation that re-emerges in the flow-matching setting.
(i)~\emph{SFT and its interactive extensions}---vanilla SFT~\citep{black2024pi0,kim2024openvla} and DAgger-style human correction~\citep{ross2011dagger}---scale to real hardware but only weakly exploit the failure signals from autonomous rollouts: vanilla SFT discards them, while DAgger uses them merely to trigger expert correction rather than as a direct optimization signal.
(ii)~\emph{Reward- or value-based RL}~\citep{ouyang2022instructgpt,schulman2017ppo,luo2024hilserl,intelligence2025pi06} requires training a reliable reward, value, or advantage model, which itself becomes a key obstacle for contact-rich manipulation where dense reward signals are difficult to obtain; HIL-SERL~\citep{luo2024hilserl} and $\pi_{0.6}*$~\citep{intelligence2025pi06} additionally introduce significant engineering overheads such as advantage values and intricate reward shaping.
(iii)~\emph{Preference-based RL} bypasses reward design via preference data: Direct Preference Optimization (DPO)~\citep{rafailov2023dpo} operates without critics by optimising likelihood ratios but was originally designed for discrete text, while recent extensions to continuous flow-matching policies, such as Flow-DPO~\citep{liu2025flowdpo} and the trajectory-level GRAPE~\citep{zhang2025grape}, restore preference learning to flow-based VLAs but inherit the reward-hacking failure mode of plain DPO and dilute the per-state learning signal.
Unlike $\pi_{0.6}*$, our \emph{FlowPRO} recipe (\ref{sec:rl_posttraining}) is entirely critic- and reward-free; unlike Flow-DPO and GRAPE, the underlying \emph{RPRO} loss anchors the implicit reward via a proximal regularizer that explicitly forbids the plain-DPO reward-hacking pathology, and exploits a contrastive-gradient-cancellation property to safely co-train on SFT samples through the same objective.

\paragraph{Asynchronous Inference and Action-Chunk Smoothing}
Action chunking~\citep{zhao2023aloha} has become the de-facto deployment recipe for VLA policies but introduces intra-chunk jitter, chunk-boundary discontinuities, and idle gaps when the backbone latency exceeds the servo period. Inference-Time RTC~\citep{pi2025rtc} introduces a lightweight flow-matching action server that refines coarse action chunks at high frequency, decoupling the slow backbone from fast control; Training-Time RTC~\citep{pi2025trainingtimertc} further co-trains this refinement module with the policy. VLASH~\citep{li2025vlash} learns an adaptive halting mechanism that determines chunk size based on task complexity, reducing inter-chunk gaps. By contrast, our deployment recipe (\ref{sec:inference}) is training-free and plug-and-play for arbitrary policies, explicitly guarantees $C^1$ continuity at chunk boundaries via tangent-aligned cubic B\'ezier curves, and is applicable to both Cartesian and joint-space control.

\section{Discussion}
\label{sec:discussion}

\paragraph{\HyVLA Pipeline}
\HyVLA co-designs data, representation, policy refinement, and deployment execution for deployable generalist robots, rather than treating the VLA as a standalone policy. Cross-embodiment deployment relies on a complete set of components beyond model scale alone: high-fidelity UMI data provides reusable supervision for learning precise manipulation priors; the compact memory encoder and rel-EE/delta-chunk representation give the policy temporal context while keeping the action interface independent of platform-specific kinematics; FlowPRO converts real failure cases into compact offline refinement without requiring large-scale online exploration; and asynchronous chunk stitching makes the same checkpoint executable under real hardware latency. These components address different bottlenecks---data quality, action representation, failure correction, and deployment timing---but they share the same goal: preserving a stable policy interface while absorbing embodiment-specific differences outside the learned core. Together, they turn \HyVLA from a single model into a practical robot-learning stack for cross-embodiment deployment.

\paragraph{Future Work}
\label{sec:limitations}
\HyVLA opens several questions that we are eager to explore, especially around data, model generalization, and real-world deployment. On the data side, an important direction is to move beyond motion capture while preserving high-precision supervision; exoskeleton-based collection is a promising route toward this goal. Since Hy-UMI-10K already provides high-accuracy action labels, it also offers a simple way to study the marginal value of precision for pre-training, for example by injecting controlled noise into the labels. In addition, the egocentric UMI camera still differs from robot-mounted deployment cameras, leaving room for systematic visual augmentation studies. To support these explorations, we will release a $2{,}000$-hour self-collected UMI subset and invite the community to study these questions and beyond.

Another key direction is real-world execution efficiency. In deployment, success is not only whether the robot can complete a task, but also whether it can execute at a practical task cadence. A key next step is therefore to improve deployment-time execution speed while maintaining safety and precision. This likely requires combining deployment-time adaptation with reinforcement learning.

Finally, the emergence of embodied intelligence remains an important open direction. \HyVLA does not study zero-shot generalization, as we believe the current data scale is still insufficient for making such claims. At the same time, recent systems such as $\pi_{0.7}$~\citep{intelligence2026pi} have begun to show early signs of zero-shot behavior, suggesting that larger-scale data and stronger pipelines may lead to qualitatively new capabilities. How to evaluate these capabilities rigorously, and how to use evaluation itself to drive the iteration of embodied models and deployment pipelines, remains an open problem.

\newpage
\renewcommand{\refname}{References}
\renewcommand{\bibname}{References}
\renewcommand{\bibsection}{\section*{\raggedright \Large References}}
\bibliographystyle{unsrtnat}
\bibliography{references_v0.5}

\newpage
\appendix
\setcounter{section}{0}
\renewcommand{\thesection}{\Alph{section}}
\begin{center}
  {\LARGE\bfseries Appendix\par}
\end{center}
\addcontentsline{toc}{section}{Appendix}
\vspace{1em}
\section{RoboTwin\,2.0 Evaluation Details}
\label{app:pertask}
\paragraph{Per-task results.} Table~\ref{tab:robotwin_pertask} reports the per-task success rates of \HyVLA on the $50$-task RoboTwin\,2.0 suite. We evaluate each task under the Clean and Randomized settings and include this breakdown as a complement to the aggregate comparison in Table~\ref{tab:robotwin_comp} (\S\ref{sec:headline_robotwin}).

\paragraph{Data filtering.} We apply an offline cleaning step because a small subset of RoboTwin\,2.0 demostrations contains implausible inverse-kinematics solutions, which often manifest as abnormal episode lengths. For each task, we cluster the episode-length distribution using HDBSCAN with cluster-selection radius $5$ and all other settings kept at defaults. This identifies stable length modes. An episode is flagged as dirty if it satisfies any of the following conditions: (i) it is assigned as an HDBSCAN noise point; (ii) it belongs to an under-populated length mode with estimated size below 100 episodes; or (iii) it lies in the top $5\%$ length tail of the longest well-populated mode. Episodes passing all three checks form the clean subset used for training.

\paragraph{Action decoding.} We let the policy predict actions under two complementary frames: (i) relative-EEF, which captures smooth local motion; and (ii) EEF, which anchors the target globally and avoids drift accumulation. The EEF based actions are concatenated after the relative ones along the chunk axis, resulting a doubled chunk size. At inference these two predictions are fused, with quaternion orientations interpolated via SLERP, combining the local precision of relative motion with the global stability of absolute targets.

\begin{table}[!htbp]
  \centering
  \footnotesize
  \setlength{\tabcolsep}{4pt}
  \renewcommand{\arraystretch}{1.05}

  \begin{tabular}{lcc@{\hskip 18pt}lcc}
    \toprule
    \textbf{Task} & \textbf{Clean} & \textbf{Rand.} & \textbf{Task} & \textbf{Clean} & \textbf{Rand.} \\
    \midrule
    adjust bottle              & 99  & 99  & place can basket           & 91  & 81  \\
    beat block hammer          & 99  & 99  & place cans plasticbox      & 100 & 100 \\
    blocks ranking rgb         & 99  & 99  & place container plate      & 98  & 99  \\
    blocks ranking size        & 93  & 94  & place dual shoes           & 94  & 95  \\
    click alarmclock           & 100 & 100 & place empty cup            & 100 & 100 \\
    click bell                 & 100 & 99  & place fan                  & 96  & 98  \\
    dump bin bigbin            & 96  & 98  & place mouse pad            & 88  & 93  \\
    grab roller                & 100 & 100 & place object basket        & 86  & 85  \\
    handover block             & 93  & 73  & place object scale         & 90  & 89  \\
    handover mic               & 87  & 98  & place object stand         & 94  & 98  \\
    hanging mug                & 37  & 33  & place phone stand          & 91  & 94  \\
    lift pot                   & 99  & 99  & place shoe                 & 98  & 100 \\
    move can pot               & 94  & 99  & press stapler              & 88  & 80  \\
    move pillbottle pad        & 96  & 96  & put bottles dustbin        & 89  & 85  \\
    move playingcard away      & 98  & 96  & put object cabinet         & 78  & 83  \\
    move stapler pad           & 94  & 95  & rotate qrcode              & 93  & 98  \\
    open laptop                & 99  & 98  & scan object                & 91  & 94  \\
    open microwave             & 72  & 64  & shake bottle               & 100 & 99  \\
    pick diverse bottles       & 88  & 72  & shake bottle horizontally  & 100 & 99  \\
    pick dual bottles          & 82  & 82  & stack blocks three         & 98  & 97  \\
    place a2b left             & 77  & 75  & stack blocks two           & 98  & 99  \\
    place a2b right            & 74  & 78  & stack bowls three          & 87  & 83  \\
    place bread basket         & 95  & 93  & stack bowls two            & 98  & 97  \\
    place bread skillet        & 94  & 90  & stamp seal                 & 85  & 81  \\
    place burger fries         & 99  & 98  & turn switch                & 50  & 49  \\
    \midrule
    \textbf{Average (50 tasks)} & \textbf{90.9} & \textbf{90.1} & & & \\
    \bottomrule
  \end{tabular}
    \vspace{4pt}
    \caption{\textbf{Per-task evaluation results of \HyVLA on the RoboTwin\,2.0 benchmark~\citep{chen2025robotwin2}.}}
    \label{tab:robotwin_pertask}
\end{table}

\section{Supplementary Deployment}

\subsection{UMI-to-Robot Deployment Derivation}
\label{app:deployment}

\paragraph{Humanoid-specific derivation.} UMI demonstrations are recorded in its own world frame and lack a torso pose. As end effector poses on Astribot S1 are defined in its own chassis frame, as mentioned in Eq.~\eqref{eq:humanoid}, and a torso pose in the chassis frame is crucial for reasonable upper-body poses and efficient IK solving, we need to find a mapping from UMI world frame to S1 chassis frame and figure out a torso pose as well. Two methodologies have been proved feasible either by our experiments or by related works:
\begin{enumerate}[leftmargin=1.6em,itemsep=2pt,topsep=2pt]
  \item \textbf{Heuristic torso/head pose inference (used in our experiments).} A lightweight rule-based estimator consumes bimanual gripper poses $\{{}^WT_{G^{\mathrm{L}}_{t}}, {}^WT_{G^{\mathrm{R}}_{t}}\}$ and infers the world-to-chassis transform, the torso pose, and the head pose such that (i)~the torso forward axis aligns with the centroid of the two gripper positions and (ii)~the torso height places both grippers within an empirically established comfortable reach shell of the upper body; see Algorithm~\ref{alg:umi2chassis} for details. It assumes that the UMI world frame and the robot chassis frame are related by a pure translation (identical orientation), which holds on Astribot S1. For robot platforms whose chassis frame has a different orientation, an additional fixed rotation can be applied without further altering the algorithm.
  \begin{algorithm}[t]
  \DontPrintSemicolon
  \SetAlgoLined
  \SetKwInOut{Input}{Input}
  \SetKwInOut{Output}{Output}
  \SetKwComment{Comment}{/* }{ */}
  \caption{Heuristic Mapping from UMI Gripper Poses to Whole-Body Targets in the Chassis Frame}
    \label{alg:umi2chassis}
    
    \Input{%
      $T^{W}_{L}, T^{W}_{R}\!\in\! SE(3)$ \Comment*[r]{UMI gripper poses in the world frame $W$}
      $L$ \Comment*[r]{full nominal arm reach (meters)}
      $h_{0}$ \Comment*[r]{nominal standing height of chassis--shoulder line}
      $\alpha\!\in\![0,1]$ \Comment*[r]{horizontal back-shift as a fraction of $L$}
      $\Delta z_{C}$ \Comment*[r]{net vertical offset for chassis localization}
      $\theta_{0}$ \Comment*[r]{constant forward torso pitch}
      $\delta\!\in\![0,1]$ \Comment*[r]{blend factor between hand height and $h_{0}$}
      $R_{\mathrm{align}}$ \Comment*[r]{fixed UMI$\rightarrow$robot gripper-axis rotation}
      $T^{T}_{H}$ \Comment*[r]{fixed torso-to-head calibration transform}
    }
    \Output{Chassis-frame targets $T^{C}_{L}, T^{C}_{R}, T^{C}_{T}, T^{C}_{H}$.}
    
    \BlankLine
    \Comment{Step 1: align UMI gripper axes to robot gripper axes}
    $T^{W}_{L} \leftarrow T^{W}_{L}\, R_{\mathrm{align}}$\;
    $T^{W}_{R} \leftarrow T^{W}_{R}\, R_{\mathrm{align}}$\;
    
    \BlankLine
    \Comment{Step 2: hand midpoint and horizontal facing direction}
    $m^{W} \leftarrow \tfrac{1}{2}\bigl(t(T^{W}_{L}) + t(T^{W}_{R})\bigr)$
      \Comment*[r]{mean of hand translations}
    $f^{W} \leftarrow \Pi_{xy}(m^{W}) \,/\, \lVert \Pi_{xy}(m^{W}) \rVert$
      \Comment*[r]{unit vector in world XY plane}
    \If{$\lVert \Pi_{xy}(m^{W}) \rVert < \varepsilon$}{
        $f^{W} \leftarrow \mathbf{e}_{x}$
          \Comment*[r]{degenerate fallback}
    }
    
    \BlankLine
    \Comment{Step 3: one-shot chassis localization (cached per episode)}
    \If{$T^{W}_{C}$ is not cached}{
        $p^{W}_{C} \leftarrow m^{W} - \alpha L\, f^{W} + \Delta z_{C}\,\mathbf{e}_{z}$
          \Comment*[r]{back-shift + vertical drop}
        $T^{W}_{C} \leftarrow (\,\mathbf{I},\, p^{W}_{C}\,)$\;
        cache $T^{W}_{C}$\;
    }
    
    \BlankLine
    \Comment{Step 4: re-express grippers and helpers in the chassis frame}
    $T^{C}_{L} \leftarrow (T^{W}_{C})^{-1}\, T^{W}_{L}$\;
    $T^{C}_{R} \leftarrow (T^{W}_{C})^{-1}\, T^{W}_{R}$\;
    $m^{C} \leftarrow (T^{W}_{C})^{-1}\, m^{W}$\;
    $f^{C} \leftarrow R(T^{W}_{C})^{\top}\, f^{W}$\;
    
    \BlankLine
    \Comment{Step 5: heuristic torso pose}
    $\psi \leftarrow \mathrm{atan2}(f^{C}_{y},\, f^{C}_{x})$
      \Comment*[r]{yaw aligned with the hands}
    $R^{C}_{T} \leftarrow R_{z}(\psi)\, R_{y}(\theta_{0})$
      \Comment*[r]{yaw, then constant forward pitch}
    $p^{C}_{T} \leftarrow \bigl(0,\, 0,\, (1-\delta)\,m^{C}_{z} + \delta\, h_{0}\bigr)^{\top}$
      \Comment*[r]{height = convex blend of hand and standing}
    $T^{C}_{T} \leftarrow (\,R^{C}_{T},\, p^{C}_{T}\,)$\;
    
    \BlankLine
    \Comment{Step 6: head by fixed torso-to-head transform}
    $T^{C}_{H} \leftarrow T^{C}_{T}\, T^{T}_{H}$\;
    
    \BlankLine
    \Return $(T^{C}_{L},\, T^{C}_{R},\, T^{C}_{T},\, T^{C}_{H})$\;
    \end{algorithm}

  \item \textbf{Whole-body IK solvers (alternative).} HoMMI-style whole-body IK~\citep{xu2026hommi} jointly resolves torso and arm configurations from EE targets and could replace the heuristic above. We document this compatibility for completeness; our Astribot\,S1 results in \S\ref{sec:headline_real_world} use the heuristic exclusively.
\end{enumerate}

\subsection{Track-B Reachability and Data Hygiene}
\label{app:reach}
Because UMI demonstrations are captured without a robot in the loop, they intrinsically lack any guarantee of reachability for arbitrary target morphologies. We deploy two standardised pre-deployment hygiene protocols, both executed \emph{offline} with zero runtime overhead:
\begin{itemize}[leftmargin=1.6em,itemsep=2pt,topsep=2pt]
  \item \textbf{Unitree G1  $\&$  Astribot\,S1.} A pre-deployment reachability verification bounds the planned task envelope to the platform; tasks exceeding the humanoid's reachable shell are excluded from this report.
  \item \textbf{JAKA K1.} The post-training UMI corpus for a given JAKA task is filtered via a single-pass IK feasibility check on the target arm; trajectories that violate JAKA's arm kinematics are removed from the post-training set.
\end{itemize}
Neither mechanism alters the policy or its action representation; they merely enforce distributional alignment between the post-training set and the physical deployment frontier.

\section{FlowPRO Hyperparameters}
\label{app:dpo}
Confirmed implementations:
\begin{itemize}[leftmargin=1.6em,itemsep=2pt,topsep=2pt]
  \item \textbf{Iterations:} $k \in \{1, 2, 3\}$; each round runs $25\,000$ optimizer steps ($75\,000$ total).
  \item \textbf{Batch size:} Global batch size of $20$ ($5$ samples per GPU).
  \item \textbf{Optimizer:} AdamW, initial learning rate $1\times 10^{-5}$, linear warmup over $1{,}000$ steps, cosine decay over the next $15{,}000$ steps to a floor of $2.5\times 10^{-6}$.
  \item \textbf{Batch composition:}
    \begin{itemize}
      \item $k=1$: $\dataset_{\text{pref}}^{1}$ / $\dataset_{\text{SFT}}$ = $80/20$.
      \item $k \geq 2$: $\dataset_{\text{pref}}^{k}$ / $\dataset_{\text{pref}}^{<k}$ / $\dataset_{\text{SFT}}$ = $70/15/15$.
    \end{itemize}
  \item \textbf{Distance metric:} To find the closest point $M'$ on $\tau^w$ for a given state $M$ on $\tau^l$, we use $d(M, M') = \| \bm{p}_M - \bm{p}_{M'} \|_2 + 0.5 \cdot d_{\text{geo}}(\bm{R}_M, \bm{R}_{M'}) + 0.2 \cdot |g_M - g_{M'}|$, where $\bm{p} \in \real^3$ is the end-effector position, $d_{\text{geo}}$ is the geodesic distance between rotation matrices in $SO(3)$, $g \in [0,1]$ is the normalized gripper width, and the weights are set empirically.
  \item \textbf{Initialization:} The Stage-2 checkpoint serves as both initialisation $\theta$ and frozen reference policy $\theta_{\mathrm{ref}}$.
  \item \textbf{Data scale:} $\leq \mathcal{O}(10^2)$ preference pairs per task; X-Trainer rollouts only.
\end{itemize}

\clearpage

\section{Author Contributions}
\label{app:authors}

\paragraph{Project supervisors.} Han Hu and Zhengyou Zhang.

\paragraph{Project leaders.} He Zhang and Lingzhu Xiang.

\paragraph{Core contributors.} Haitao Lin, Zeyu Huang, Minghui Wang, Dingyan Zhong, Yubo Dong, Yihao Wu, and Yongming Rao.

\paragraph{Contributors.} Dongsheng Zhang, Wanjia He, Ling Chen, Kai Huang, Jiahao Chen, Sichang Su, Xumin Yu, Ziyi Wang, Chengwei Zhu, Xiao Teng, Yuchun Guo, Yufeng Zhang, Yuandong Liu, Rui Wang and Zisheng Lu.

\end{document}